\documentclass[letterpaper]{article} % DO NOT CHANGE THIS
\usepackage{aaai23}  % DO NOT CHANGE THIS
\usepackage{times}  % DO NOT CHANGE THIS
\usepackage{helvet}  % DO NOT CHANGE THIS
\usepackage{courier}  % DO NOT CHANGE THIS
\usepackage[hyphens]{url}  % DO NOT CHANGE THIS
\usepackage{graphicx} % DO NOT CHANGE THIS
\usepackage{natbib}  % DO NOT CHANGE THIS AND DO NOT ADD ANY OPTIONS TO IT
\usepackage{caption} % DO NOT CHANGE THIS AND DO NOT ADD ANY OPTIONS TO IT
\usepackage{algorithm}
\usepackage{algorithmic}
\usepackage{subfig}
\usepackage{graphicx}
\usepackage{url}            % simple URL typesetting
\usepackage{booktabs}       % professional-quality tables
\usepackage{amsfonts}       % blackboard math symbols
\usepackage{nicefrac}       % compact symbols for 1/2, etc.
\usepackage{microtype}      % microtypography
\usepackage{xcolor}         % colors
\usepackage{multirow}
\usepackage{textcomp, gensymb}
\usepackage{amsmath}
\usepackage{subfig}
\usepackage{tikz}
\usepackage{bigdelim}
\usepackage{graphicx}
\usepackage{newfloat}
\usepackage{listings}
\DeclareCaptionStyle{ruled}{labelfont=normalfont,labelsep=colon,strut=off} % DO NOT CHANGE THIS
\floatstyle{ruled}
\newfloat{listing}{tb}{lst}{}
\floatname{listing}{Listing}
\newcommand{\tblak}[1]{\textcolor{black}{ #1 }}

\newcommand{\our}{CrysGNN}

\title{CrysGNN : Distilling pre-trained knowledge to enhance property prediction for crystalline materials.}
\author {
    Kishalay Das,\textsuperscript{\rm 1}
    Bidisha Samanta,\textsuperscript{\rm 1}
    Pawan Goyal,\textsuperscript{\rm 1}
    Seung-Cheol Lee,\textsuperscript{\rm 2}\\
    Satadeep Bhattacharjee,\textsuperscript{\rm 2}
    Niloy Ganguly\textsuperscript{\rm 1,3}
}
\affiliations {
    \textsuperscript{\rm 1} Indian Institute of Technology Kharagpur, India
    \textsuperscript{\rm 2} Indo Korea Science and Technology Center, Bangalore, India.\\
    \textsuperscript{\rm 3} L3S, Leibniz University of Hannover, Germany.\\
    kishalaydas@kgpian.iitkgp.ac.in, bidisha@iitkgp.ac.in, pawang@cse.iitkgp.ac.in\\
    seungcheol.lee@ikst.res.in, s.bhattacharjee@ikst.res.in, niloy@cse.iitkgp.ac.in
}

\begin{document}

\maketitle
\begin{abstract}
In recent years, graph neural network (GNN) based approaches have emerged as a powerful technique to encode complex topological structure of crystal materials in an enriched representation space. These models are often supervised in nature and using the property-specific training data, learn relationship between crystal structure and different properties like formation energy, bandgap, bulk modulus, etc. Most of these methods require a huge amount of property-tagged data to train the system which may not be available for different properties. However, there is an availability of a huge amount of crystal data with its chemical composition and  structural bonds. To leverage  these untapped data, this paper presents \our{}, a new pre-trained GNN framework for crystalline materials, which captures both node and graph level structural information of crystal graphs using a huge amount of unlabelled material data. Further, we extract distilled knowledge from \our{} and inject into different state of the art property predictors to enhance their property prediction accuracy. We conduct extensive experiments to show that with distilled knowledge from the pre-trained model, all the SOTA algorithms are able to outperform their own vanilla version with good margins.  We also observe that the distillation process provides a significant improvement over the conventional approach of finetuning the pre-trained model. We have released
% \footnote{Source code, pre-trained model, and dataset of CrysGNN is made available at https://github.com/kdmsit/crysgnn} 
the pre-trained model along with the large dataset of 800K crystal graph which we carefully curated;  so that the pre-trained model can be plugged into any existing and  upcoming models to enhance their prediction accuracy.
\end{abstract}
\section{Introduction}
\label{intro}
Fast and accurate prediction of different material properties is a challenging and important task in material science.  In recent times there has been an ample amount of data driven works ~\cite{seko2015prediction,pilania2015structure,lee2016prediction,de2016statistical,seko2017representation,isayev2017universal,ward2017including,lu2018accelerated,im2019identifying}
for predicting crystal properties which are as accurate as theoretical DFT [Density functional Theory] based approaches~\cite{orio2009density}, however, much faster than it. The architectural innovations of these approaches towards accurate property predictions come from incorporating specific domain knowledge into a deep encoding module. For example, in order to encode the neighbourhood structural information around a node (atom), GNN based approaches ~\cite{xie2018crystal,chen2019graph,louis2020graph,Wolverton2020,schmidt2021crystal} gained some popularity in this domain. Understanding the importance of many-body interactions, ALIGNN~\cite{choudhary2021atomistic} incorporates bond angular information into their encoder module and became SOTA for a large range of property predictions. However, as different properties expressed by a  crystalline material are a complex function of different inherent structural and chemical properties of the constituent atoms, it is extremely difficult to explicitly incorporate them into the encoder architecture. Moreover, data sparsity across properties is a known issue~\cite{das2022crysxpp,jha2019enhancing}, which makes these models difficult to train for all the properties. To circumvent this problem we adopt the concept of self-supervised pre-training \cite{devlin2018bert,trinh2019selfie,chen2020simple,chen2020adversarial,he2020momentum,hu2020pretraining,hu2020gpt,qiu2020gcc,you2020graph} for crystalline materials which enables us to leverage a large amount of untagged material structures to learn the complex hidden features which otherwise are difficult to identify.\\
In this paper, we introduce a graph pre-training method which captures (a) connectivity of different atoms, (b) different atomic properties and (c) graph similarity from a large set of unlabeled data. To this effect, we curate a new large untagged crystal dataset with 800K crystal graphs and undertake a  pre-training framework (named  \our{}) with the dataset. \our{}  learns the representation of a crystal graph by initiating self-supervised loss  at both node (atom) and graph (crystal) level. At the node level, we pre-train the GNN model to reconstruct the node features and connectivity between nodes in a self-supervised way, whereas at the graph level, we adopt supervised and contrastive learning to learn structural similarities between graph structures using the space group and crystal system information of the crystal materials respectively.\\
We subsequently distill important structural and chemical information of a crystal from the pre-trained \our{} model and pass it to the property predictor. 
The distillation process  provides wider usage than the conventional pretrain-finetuning framework as transferring pre-trained knowledge to a property predictor and finetuning it requires a similar graph encoder architecture between the pre-trained model and the property predictor, which limits the knowledge transfer capability of the pre-trained model. On the other hand, using knowledge distillation \cite{romero2014fitnets,hinton2015distilling}, we can retrofit the pre-trained \our{} model into any existing state-of-the-art property predictor, irrespective of their architectural design, to improve their property prediction performance. Also experimental results (presented later) show that even in case of similar graph encoder, distillation performs better than finetuning.\\ 
% By design, this knowledge distillation based approach is more robust and independent of the underlying architecture of the property predictor,thus it can enhance the performance of a diverse set of SOTA models
% This makes our pre-trained \our{} model more generic and versatile.
%\noteng{needs to be shortened}\\
With rigorous experimentation %on 19 different properties 
across two popular benchmark materials datasets, we show that distilling necessary information from \our{}
\footnote{Source code, pre-trained model, and dataset of CrysGNN is made available at https://github.com/kdmsit/crysgnn} 
to various property predictors results in substantial performance gains for  GNN based architectures and  complex ALIGNN model. The improvements range from 4.19\% to 16.20\% over several highly optimized SOTA models. We also perform ablation studies to investigate the influence of different pre-training losses in enhancing the SOTA model performance and observe significant performance gain employing the both node and graph-level pre-training, compared to node-level or graph-level pre-training separately. Also using both supervised and contrastive graph-level pre-training, we are able to learn more robust and expressive graph representation which enhances the property predictor performance.
This also helps to achieve even better improvements when the dataset is sparse. Moreover, the property-tagged dataset suffers from certain biases as it is theoretically (DFT) derived, hence the property predictor also suffers from such bias.  We found that on being trained with small amount of experimental data, the DFT bias decreases substantially.

\section{Related Work}
\label{related_work}
In recent times, data driven approaches~\cite{seko2015prediction,pilania2015structure,lee2016prediction,de2016statistical,seko2017representation,isayev2017universal,ward2017including,lu2018accelerated,im2019identifying} have become quite popular to establish relationship between the atomic structure of crystalline materials and their properties with very high precision. Especially, graph neural network (GNN) based approaches  ~\cite{xie2018crystal,chen2019graph,louis2020graph,Wolverton2020,schmidt2021crystal,choudhary2021atomistic} have emerged as a powerful machine learning model tool to encode material’s complex topological structure along with  node features in an enriched representation space.\\
% Models such as CGCNN~\cite{xie2018crystal}, GATGNN~\cite{louis2020graph} represent 3D crystal structure as a multi-graph and build a graph convolution neural network directly on the graph to update node features based on their local chemical and structural environment. MEGNet~\cite{chen2019graph} introduces global state attributes for quantitative structure-state property relationship prediction in materials, whereas ALIGNN~\cite{choudhary2021atomistic} explicitly captures many body interactions by incorporating bond angles and local geometric distortions. \\
There are attempts to pre-train GNNs to extract graph and node-level representations.~\cite{hu2020pretraining} develops an effective pre-training strategy for GNNs, where they perform both node-level and graph-level pre-training on GNNs to capture domain specific knowledge about nodes and edges, in addition to global graph-level knowledge. Followed by this work, there has been several other works on self-supervised graph pre-training ~\cite{hu2020gpt,qiu2020gcc,you2020graph},  which propose different graph augmentation methods and maximizes the agreement between two augmented views of the same graph via a contrastive loss. In the field of crystal graphs, CrysXPP~\cite{das2022crysxpp} is the only model which comes close to a pre-trained model. In their work, an autoencoder is trained on  a  volume of un-tagged crystal graphs and  the learned knowledge is (transferred to)  used to initialize the encoder of  CrysXPP, which is fine-tuned with property specific tagged data. \\
\tblak{Although conceptually similar to the work done by ~\citet{hu2020pretraining}, our work differs in the following three key aspects: (1) pre-training strategy proposed by ~\citeauthor{hu2020pretraining} is very effective for molecular dataset, but it is difficult to extend directly to crystalline material because structural semantics are different between molecules and materials \cite{xie2021crystal}. 
Molecules have non-periodic and finite structures,  solid materials’ structures are infinite and periodic in nature.
% Moreover, a molecule typically contains 5-10 atom types or elements (Like $N$, $H$, $O$), but in materials, we can have any of the 94 naturally occurring elements in the periodic table.
(2) For graph-level pre-training, ~\citeauthor{hu2020pretraining} adapted supervised graph-level property prediction using a huge amount of labelled dataset from chemistry and biology domain, which makes it less effective in several other domains like material science where property labeled data is extremely scarce. Also, a crucial step in graph-level prediction is to find graph structural similarity between two sets of graphs, which they do not explore but mention as a future work. We do not make use of supervised pre-training which requires a large amount of property tagged material data. Instead, we leverage the idea of structural similarity of materials belonging to the similar space group, and via contrastive loss and space group classification loss, we try to capture this similarity. (3) Finally they follow conventional pre-train finetuning framework, whereas in \our{}  we incorporate the idea of knowledge distillation \cite{romero2014fitnets,hinton2015distilling} to distill important information from the pre-trained model and inject it into the property prediction process. By design, this knowledge distillation based approach is more robust and independent of the underlying architecture of the property predictor, thus it can enhance the performance of a diverse set of SOTA models.}
\section{Methodology}
\label{methodology}
\begin{figure*}
	\centering
	\vspace*{-1mm}
	\subfloat[Node-level decoding]{
		\boxed{\includegraphics[width=\columnwidth, height=40mm]{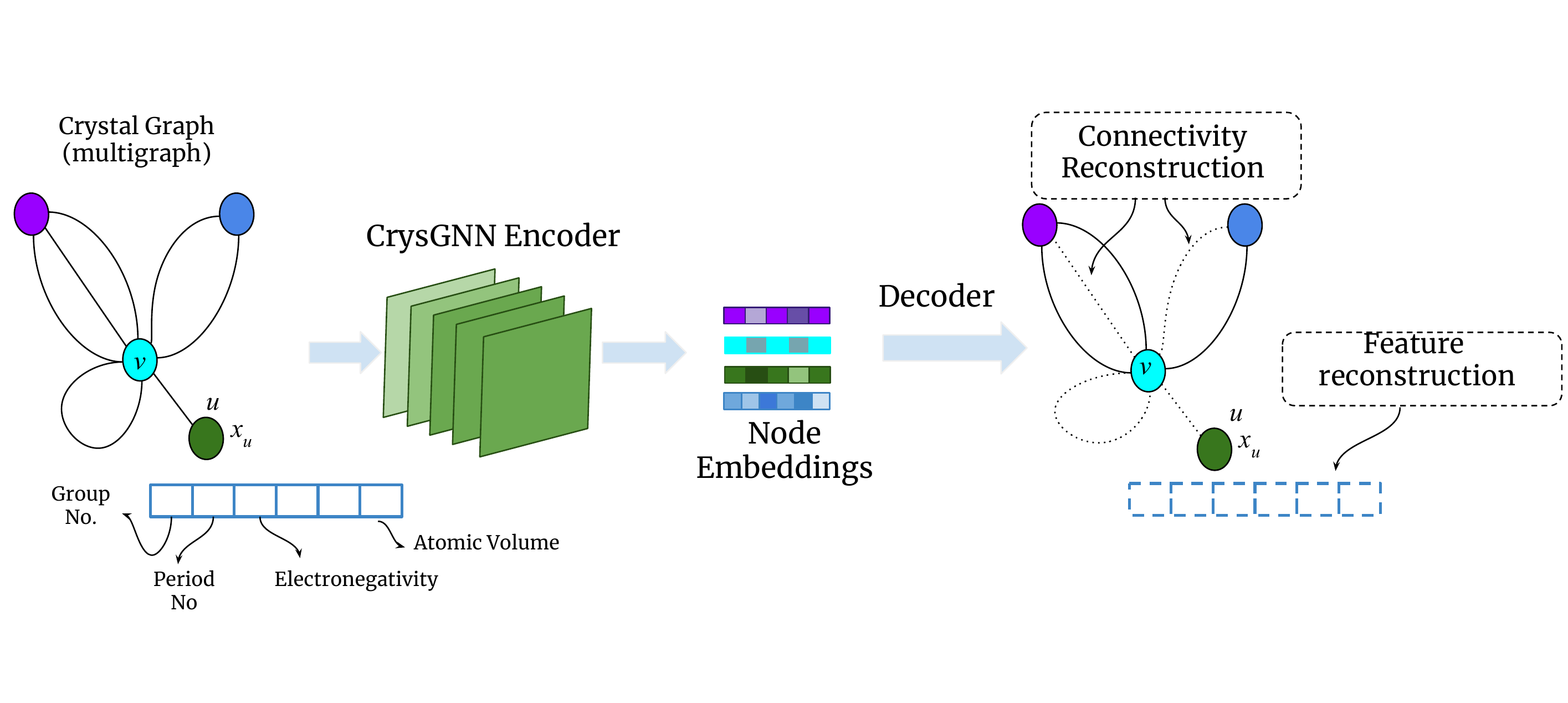}}}
	\subfloat[Graph-level decoding]{
		\boxed{\includegraphics[width=\columnwidth, height=40mm]{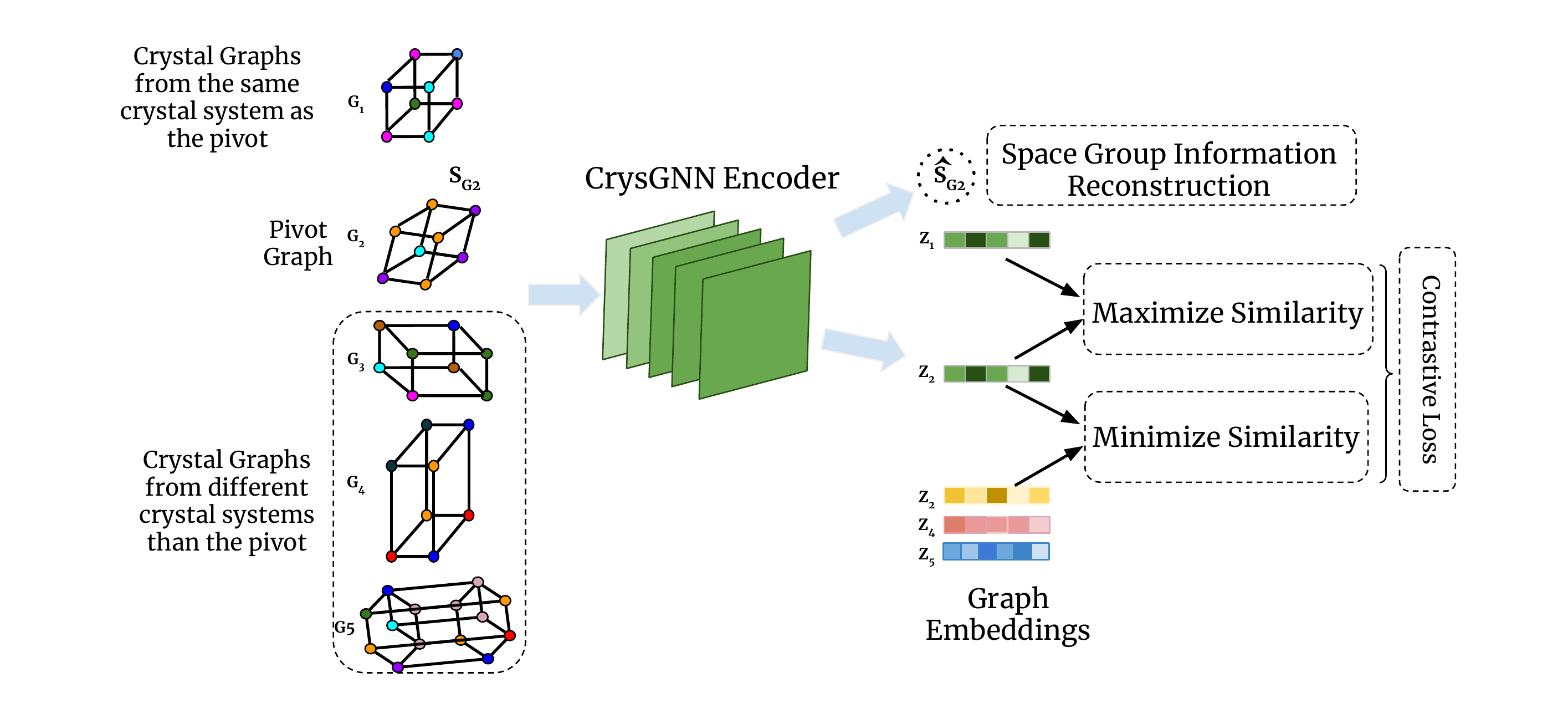}}}
	\caption{Overview of both node and graph-level decoding methods for \our{}. (a) In node-level decoding, node feature attributes and connectivity between nodes are reconstructed in a self-supervised way. (b) In graph-level decoding, $G_2$ is the pivot graph and $G_1$ is from the same crystal system (Cubic), whereas $G_3, G_4, G_5$ are from different crystal systems. First we reconstruct space group information of $G_2$, then through contrastive loss, \our{} will maximize similarities between positive pair ($G_2,G_1$) and minimize similarities between negative pairs ($G_2, G_3$), ($G_2, G_4$) and ($G_2, G_5$) in embedding space.}% \pg{Description in 3.1.2 is slightly different, you talk about a positive pair and k negative pairs per G\_i. The figure/text may have to be changed to fit that.}}
	\label{fig:CrysGNN}
\end{figure*}
Formally, we first curate a huge amount of property un-tagged crystal graphs $\mathcal{D}_{u} = \{\mathcal{G}_i\}$ from various materials datasets to pre-train a deep GNN model $f_{\theta}$, that learns intrinsic structural and chemical patterns of the crystal graphs.
Further, we use a training set of property tagged crystal graphs $\mathcal{D}_{t} =\{\mathcal{G}_i,y_i\}$ for property prediction, which is smaller in volume and \tblak{may or may not be disjoint from the original untagged set} $\mathcal{D}_{u}$. We train any supervised property predictor $\mathcal{P}_{\psi}$ using $\mathcal{D}_{t}$ to predict the property value given the crystal graph structure. While training the property predictor, we incorporate the idea of knowledge distillation to distill important structural and chemical information from the pre-trained model. This knowledge may prove to  be useful to a property predictor which now need not learn from scratch, but be armed with distilled  knowledge from the pre-trained model.  Hence in this section, we first describe the \our{} pre-training strategy, followed by the knowledge distillation and property prediction process. %\pg{Section references need to be modified / removed as AAAI does not have section numbers by default.}
\subsection{\our{} Pre-training} 
\label{pretraining}
We build a deep auto-encoder architecture \our{}, which comprises a graph convolution based encoder followed by an effective decoder. The autoencoder is (pre)trained end to end, using a large amount of property un-tagged crystal graphs $\mathcal{D}_{u} = \{\mathcal{G}_i\}$, where via node and graph-level self-supervised losses, the model can capture the structural and chemical information of the crystal graph data. First, we formalize the representation of a crystal 3D structure into a multi-graph structure, which will be an input to the encoder module.
\subsubsection{Crystal Graph Representation.}
\label{graph_rep}
We realize a crystal material as a multi-graph structure $\mathcal{G}_i =(\mathcal{V}_i, \mathcal{E}_i, \mathcal{X}_i, \mathcal{F}_i )$ as proposed in \cite{xie2018crystal}. $\mathcal{G}_i$ is an undirected weighted multi-graph where $\mathcal{V}_i$ denotes the set of nodes or atoms present in a unit cell of the crystal structure. $\mathcal{E}_i=\{(u,v,k_{uv})\}$ denotes a multi-set of node pairs and $k_{uv}$ denotes number of edges between a node pair $(u,v)$. $\mathcal{X}_i=\{(x_{u} | u \in \mathcal{V}_i )\}$ denotes the node feature set proposed by CGCNN~\cite{xie2018crystal}. It includes different chemical properties like electronegativity, valance electron, covalent radius, etc. Finally, $\mathcal{F}_i=\{\{s^k\}_{(u,v)} |  (u,v) \in \mathcal{E}_i, k\in\{1..k_{uv}\}\}$ denotes the multi-set of edge weights where $s^k$ corresponds to the $k^{th}$ bond length between a node pair $(u,v)$, which signifies the inter-atomic bond distance between two atoms. Next, we formally define \our{} pre-training  and knowledge distillation based property prediction strategy.
\begin{figure*}
	\centering
	\vspace*{-1mm}
	\boxed{\includegraphics[width=1.3\columnwidth]{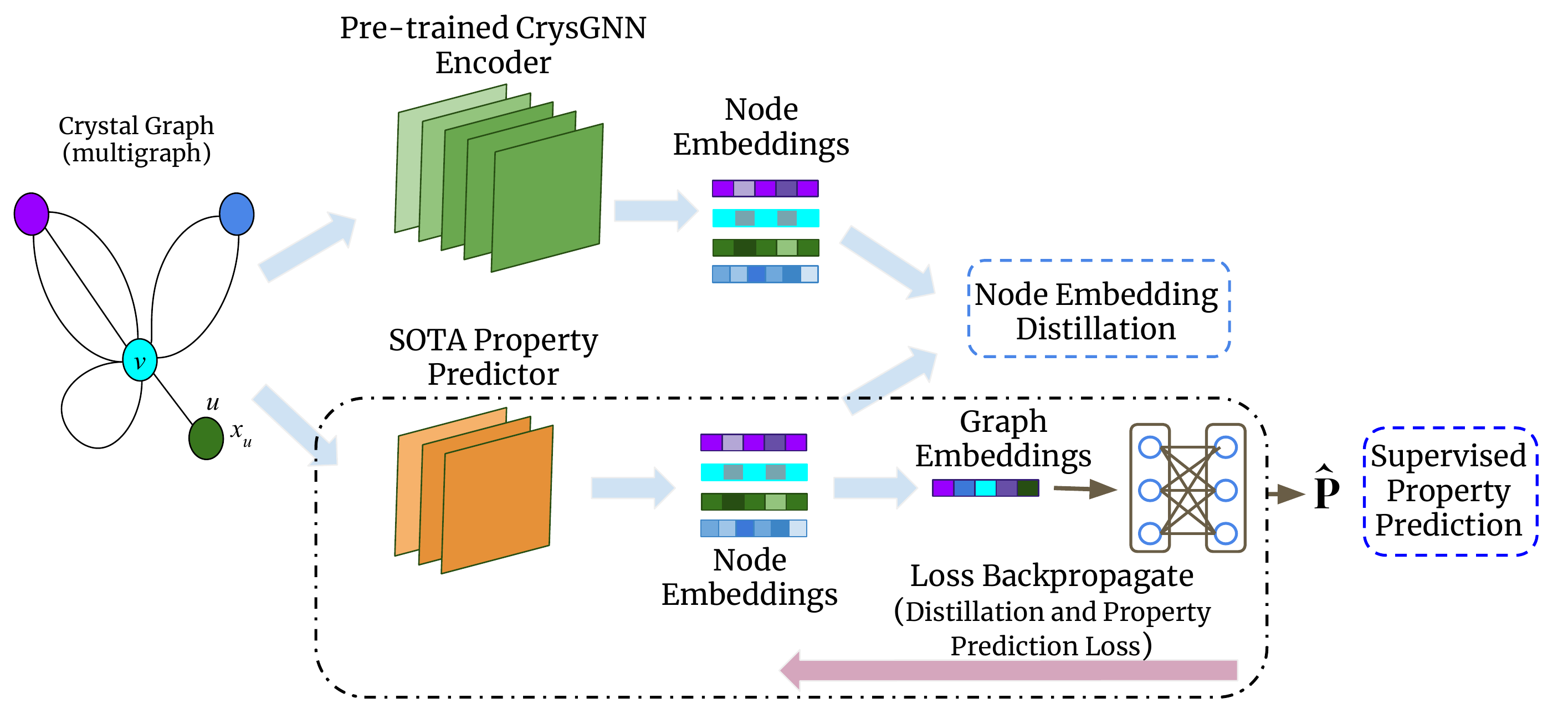}}
	\caption{Overview of Property Prediction using Knowledge Distillation from \our{}.}
	\label{fig:dist_prop}
\end{figure*}
\subsubsection{Self Supervision.}
\label{self_sup}
We first develop a graph convolution  \cite{xie2018crystal} based encoding module, which takes crystal multi-graph structure $\mathcal{G} =(\mathcal{V}, \mathcal{E}, \mathcal{X}, \mathcal{F} )$ as input and encodes structural semantics of the crystal graph into lower dimensional space. Each layer of convolution, follows an iterative neighbourhood aggregation (or message passing) scheme to capture the structural information within node's (atom's) neighbourhood. After $L$-layers of such aggregation, the encoder returns the final set of node embeddings $\mathcal{Z} = \{z_1,..., z_{|\mathcal{V}|}\}$, where $z_u :=z_u^L$ represents the final embedding of node $u$.  Details of the GNN architecture is in Appendix.
Next, we design an effective decoding module, which takes node embeddings $\mathcal{Z}$ as input and learns local chemical features and global structural information through node and graph-level decoding, respectively. Decoding node-level information will enable \our{} to learn local domain specific chemical features and connectivity information around an atom, while decoding graph-level features  will help \our{} capture global structural knowledge. \\\\
\textbf{Node-Level Decoding.} For node-level decoding (Fig-\ref{fig:CrysGNN}(a)), we propose two self-supervised learning methods, where we reconstruct two important features that induce the local chemical environment of the crystal around a node (atom). For a given node $u$, we first reconstruct its node features $x_u$, %\pg{Your earlier notation for features is different and this particular terminology is not coined earlier} 
which represent different chemical properties of atom $u$. Given a node embedding $z_u$, which is encoded based on neighbouring structure around atom $u$, we apply a linear transformation on top of $z_u$ to reconstruct the node attributes. In crystalline graphs, node features correspond to different chemical properties associated with the constituent atoms 
% like atom type, its electronegativity, valance electron, electron affinity, atomic volume, etc. 
through reconstructing these features \our{}  captures local chemical semantics around that atom.\\ %\pg{You have already mentioned this earlier in Crystal Graph Representation. }
Further, we reconstruct local connectivity around an atom, where given node embeddings of two nodes $u$ and $v$, we apply a bi-linear transformation module to generate combined transformed embedding of two nodes $z_{uv}$, which we pass through a feed forward network to predict the strength of association between two atoms. Through reconstructing this local connectivity around an atom, \our{}  encodes the periodicity of the node i.e. the number of neighbours around it along with the relative position of its neighbours and their bond length.
\\\\
\textbf{Graph-level Decoding.} 
We aim to capture periodic structure of a crystal material through graph-level decoding (Fig-\ref{fig:CrysGNN}(b)). We specifically leverage two concepts in doing so. (a). {\bf Space group} which is used to describe the symmetry of a unit cell of the crystal material. In materials science literature there are  230 unique space groups and each crystal (graph) has a unique space group number. 
(b). {\bf Crystal system.} The space group level information can classify a crystal graph into 7 broad groups of crystal systems like Triclinic, Monoclinic, Orthorhombic, Tetragonal, Trigonal, Hexagonal, and Cubic. Several electronic and optical properties such as band gap, dielectric constant depend on the space-group and the crystal structure justifying its usage. More information about different space groups and crystal systems is in Appendix

%where we leverage space group 
% ~\footnote{Space group is defined as a group of symmetry operations that are specifies different ways a crystal can be oriented without changing its atoms' position. In materials science literature there are  230 unique space groups. Moreover, space group level information can classify a crystal graph into 7 broad group of crystal systems like Triclinic, Monoclinic, Orthorhombic, Tetragonal, Trigonal, Hexagonal, and Cubic.} 
%~\footnote{Space group is defined as a group of symmetry operations that are combined to describe the symmetry of a region in 3-dimensional space, the unit cell of the crystal material. It specifies different ways a crystal can be oriented without changing its atoms' position. In materials science literature there are  230 unique space groups. Moreover, space group level information can classify a crystal graph into 7 broad group of crystal systems like Triclinic, Monoclinic, Orthorhombic, Tetragonal, Trigonal, Hexagonal, and Cubic.} 
%level information of the underlying crystal graph through supervised and self supervised learning methods.\\

\tblak{Given the set of node embeddings $\mathcal{Z} = \{z_1,..., z_{|\mathcal{V}|}\}$, we use a symmetric aggregation function to generate graph-level representation $\mathcal{Z}_{\mathcal{G}}$. First, we pass $\mathcal{Z}_{\mathcal{G}}$ through a feed-forward neural network to predict the space group number of graph $\mathcal{G}$.
Further, we develop a contrastive learning framework for pre-training of \our{}, where pre-training is performed by maximizing (minimizing) similarity between two crystal graphs belonging to the same (different) crystal system via contrastive loss in graph embedding space.}  %Lattice Structure of crystalline materials will be similar for the materials belonging to the same crystal system and change across different crystal systems. Moreover, electronic and optical properties such as band gap, dielectric constant or transport properties such as Seebeck coefficients depend on the spacegroup or more precisely the details of the crystal structure. Hence capturing graph-level semantics is a useful pretraining task and we use contrastive loss for that.} \noteng{Is crystal system means space group?}
A mini-batch of $N$ crystal graphs is randomly sampled and processed through contrastive learning to 
align the positive pairs $\mathcal{Z}_{\mathcal{G}_i},\mathcal{Z}_{\mathcal{G}_j}$ of graph embeddings, which belong to the same crystal system and contrast the negative pairs which are from different crystal systems. Here we adopt the
normalized temperature-scaled cross-entropy loss (NT-Xent)\cite{sohn2016improved,van2018representation,wu2018unsupervised} and NT-Xent for the $i^{th}$ graph is defined:
\begin{equation}
    \label{eq:contrastive}
        % \mathcal{L}_{NTXent}
        \mathcal{L}_{i}=-log\frac{exp(sim(\mathcal{Z}_{\mathcal{G}_i},\mathcal{Z}_{\mathcal{G}_j})/\tau)}{\sum_{k=1}^{K} exp(sim(\mathcal{Z}_{\mathcal{G}_i},\mathcal{Z}_{\mathcal{G}_k})/\tau)}
\end{equation}
where $\tau$ denotes the temperature parameter and  $sim(\mathcal{Z}_{\mathcal{G}_i},\mathcal{Z}_{\mathcal{G}_j})$ denotes cosine similarity function.
\tblak{The final loss $\mathcal{L}_{NTXent}$ is computed across all positive pairs in the minibatch.} 
Overall we pre-train this deep auto-encoder architecture \our{} end to end to optimize the following loss :
\begin{equation}
    \label{eq:pretrain_loss}
        \mathcal{L}_{pretrain}=\alpha\mathcal{L}_{FR}+\beta\mathcal{L}_{CR}+\gamma\mathcal{L}_{SG}+\lambda\mathcal{L}_{NTXent}
\end{equation}
where $\mathcal{L}_{FR},\mathcal{L}_{CR}$ are the reconstruction losses for node feature, and local connectivity, $,\mathcal{L}_{SG}$ is the space group supervision loss, $\mathcal{L}_{NTXent}$ is the contrastive loss and $\alpha$, $\beta$, $\gamma$, $\lambda$ are the weighting coefficients of each loss. We denote the set of parameters in \our{} model as $\theta$ and the pre-trained \our{} as $f_{\theta}$.
\subsection{Distillation and Property Prediction}
We aim to retrofit the pre-trained \our{} model into any SOTA property predictor to enhance its learning process and improve performance (Fig-\ref{fig:dist_prop}). Hence we incorporate the idea of knowledge distillation to distill important structural and chemical information from the pre-trained model, which is useful for the downstream property prediction task, and feed it into the property prediction process.
% With the pre-trained model, our goal is to design a property predictor (Figure \ref{fig:dist_prop}), which will be trained using the property specific tagged dataset $\mathcal{D}_{t} =\{\mathcal{G}_i,y_i\}$.
% We do not finetune the pretrained \our{} on specific property, because it is difficult to learn optimal parameter space with limited data and may suffer from negative transfer. 
% \tblu{We aim to
% retrofit the pre-trained model into any SOTA property predictor to enhance its learning process and improve performance.} 
Formally, given the pre-trained \our{} model $f_{\theta}$, any SOTA property predictor $\mathcal{P}_{\psi}$ and set of property tagged training data $\mathcal{D}_{t} =\{\mathcal{G}_i,y_i\}$, we aim to find optimal parameter values ${\psi^*}$ for $\mathcal{P}$. We train $\mathcal{P}_{\psi}$ using dataset $\mathcal{D}_{t}$ to optimize the following multitask loss:
\begin{equation}
    \label{eq:finetune_loss}
        \mathcal{L}_{prop}=\delta\mathcal{L}_{MSE}+(1-\delta)\mathcal{L}_{KD}
\end{equation}
where $\mathcal{L}_{MSE} = (\hat{y_i}-y_i)^2$ denotes the discrepancy between predicted and true property values by $\mathcal{P}_{\psi}$ (property prediction loss). We define knowledge distillation loss $\mathcal{L}_{KD}$ to match intermediate node feature representation between the pre-trained \our{} model and the SOTA property predictor $\mathcal{P}_{\psi}$ as follows:
\begin{equation}
\label{eq:kd_loss}
                \mathcal{L}_{KD}=  \lVert {\mathcal{Z}^{T}_{i}} - {\mathcal{Z}^{S}_{i}} \rVert^2 
\end{equation}
where $\mathcal{Z}^{T}_{i}$ and $\mathcal{Z}^{S}_{i}$ denote intermediate node embeddings of the pre-trained \our{} and the property predictor $\mathcal{P}_{\psi}$ for crystal graph $\mathcal{G}_i$, respectively. \tblak{Note, both $\mathcal{Z}^{T}_{i}$ and $\mathcal{Z}^{S}_{i}$ are projected on the same latent space.} Finally, $\delta$ signifies relative weightage between two losses, which is a hyper-parameter to be tuned on validation data. During property prediction the pre-trained network is frozen and we backpropagate $\mathcal{L}_{prop}$ through the predictor $\mathcal{P}_{\psi}$ end to end.
\begin{table}
  \centering
  \small
    \setlength{\tabcolsep}{2 pt}
    \scalebox{0.68}{
    \begin{tabular}{c | c | c | c | c | c }
    \toprule
    Task & Datasets & Graph Num. & Structural Info.  & Properties Count & Data Type\\
    \midrule
     \multirow{2}{*}{\shortstack{Pre-training}}
     & OQMD & 661K & \checkmark  & x & DFT Calculated\\
     & MP & 139K & \checkmark  & x & DFT Calculated\\
     \midrule
    \multirow{3}{*}{\shortstack{Property \\ Prediction}} & MP 2018.6.1 & 69K & \checkmark  & 2 & DFT Calculated\\
     & JARVIS-DFT & 55K & \checkmark  & 19 & DFT Calculated\\
     & OQMD-EXP & 1.5K & \checkmark  & 1 & Experimental\\
    \bottomrule
  \end{tabular}
  }
  \caption{Datasets Details}
  \label{tbl-dataset}
\end{table}
\begin{table*}
  \centering
  \small
    \setlength{\tabcolsep}{11 pt}
    \scalebox{0.9}{
      \begin{tabular}{c | c c | c c| c c | c c}
        \toprule
        Property &  CGCNN & CGCNN & CrysXPP & CrysXPP & GATGNN & GATGNN & ALIGNN & ALIGNN \\
        & & (Distilled) &  & (Distilled) &  & (Distilled) &  & (Distilled)\\
        \midrule
        Formation Energy & 0.039 & \textbf{0.032} & 0.041 & \textbf{0.035} & 0.096 & \textbf{0.091} & 0.026 & \textbf{0.024}  \\
        Bandgap (OPT)     & 0.388 & \textbf{0.293} & 0.347 & \textbf{0.287} & 0.427 & \textbf{0.403} & 0.271 & \textbf{0.253}  \\
        \midrule
        Formation Energy  & 0.063 & \textbf{0.047} & 0.062 & \textbf{0.048} & 0.132 & \textbf{0.117} & 0.036 & \textbf{0.035}  \\
        Bandgap (OPT)     & 0.200 & \textbf{0.160} & 0.190 & \textbf{0.176} & 0.275 & \textbf{0.235} & 0.148 & \textbf{0.131} \\
        Total Energy     & 0.078 & \textbf{0.053} & 0.072 & \textbf{0.055} & 0.194 & \textbf{0.137} & 0.039 & \textbf{0.038}  \\
        Ehull             & 0.170 & \textbf{0.121} & 0.139 & \textbf{0.114} & 0.241 & \textbf{0.203} & 0.091 & \textbf{0.083} \\
        Bandgap (MBJ)     & 0.410 & \textbf{0.340} & 0.378 & \textbf{0.350} & 0.395 & \textbf{0.386} & 0.331 & \textbf{0.325} \\
        Spillage     & 0.386 & \textbf{0.374} & 0.363 & \textbf{0.357} & 0.350 & \textbf{0.348} & 0.358 & \textbf{0.356} \\
        SLME (\%)     & 5.040 & \textbf{4.790} & 5.110 & \textbf{4.630} & 5.050 & \textbf{4.950} & 4.650 & \textbf{4.590} \\
        Bulk Modulus (Kv)     & 12.45 & \textbf{12.31} & 13.61 & \textbf{12.70} & 11.64 & \textbf{11.53} & 11.20 & \textbf{10.99} \\
        Shear Modulus (Gv)     & 11.24 & \textbf{10.87} & 11.20 & \textbf{10.56} & 10.41 & \textbf{10.35} & 9.860 &\textbf{9.800}  \\
         \bottomrule
      \end{tabular} 
      }
   \caption{Summary of the prediction performance (MAE) of different properties in Materials project (Top)  and JARVIS-DFT (Bottom). Model M is the vanilla variant of a SOTA model and M  (Distilled) is the distilled variant using the pretrained \our{}. The best performance is highlighted in bold.}
  \label{tbl-full-data-1}
\end{table*}

\section{Experimental Results}
\label{exp_results}
In this section, we evaluate how the distilled knowledge from \our{} enhances the performance of different state of the art property predictors on a diverse set of crystal properties from two popular benchmark materials datasets. We first briefly discuss the datasets used both in pre-training and downstream property prediction tasks. Then we report the results of different SOTA property predictors on the downstream property prediction tasks. 
Next, we illustrate the effectiveness of our knowledge distillation method compared to the conventional fine-tuning approach.
We further conduct some ablation studies to show the influence of different pre-training losses in predicting different crystal properties and the performance of the system to sparse dataset. Finally, we demonstrate how distilled knowledge from the pre-trained model aids the SOTA models to remove DFT error bias, using very little experimental data.
\subsection{Datasets}
We curated 800K untagged crystal graph data from two popular materials databases, Materials Project (MP) and OQMD, to pre-train \our{} model. Further to evaluate the performance of different SOTA models with distilled knowledge from \our{}, we select MP 2018.6.1 version of Materials Project and  2021.8.18 version of JARVIS-DFT, another popular materials database, for property prediction as suggested by \cite{choudhary2021atomistic}.  \tblak{Please note, MP 2018.6.1 dataset is a subset of the dataset used for pre-training, whereas JARVIS-DFT is a separate dataset which is not seen during the pre-training.}
MP 2018.6.1 consists of 69,239 materials with two properties bandgap and formation energy, whereas JARVIS-DFT consists of 55,722 materials with 19  properties \tblak{which can be broadly classified into two categories : 1) properties like formation energy, bandgap, total energy, bulk modulus, etc. which depend greatly on crystal structures and atom features, and 2) properties like $\epsilon_x$, $\epsilon_y$, $\epsilon_z$, n-Seebeck, n-PF, etc. which depend on the precise description of the materials’ electronic structure. In the following section, we will evaluate effectiveness of \our{} on the first class of properties. The impact of structural information is marginal on the  second class of property hence all the SOTA perform poorly, there is however some modest improvement using \our{}; we have put the results  and the discussion about it in the Appendix.}\\
Moreover, all these properties in both Materials Project and JARVIS-DFT datasets are based on DFT calculations of chemicals. Therefore, to investigate how pre-trained knowledge helps to mitigate the DFT error, we also take a small dataset OQMD-EXP~\cite{kirklin2015open}, containing 1,500 available experimental data of formation energy. Details of each of these datasets are given in Table \ref{tbl-dataset}. More detail about dataset, different crystal properties and experimental setup is in Appendix
\subsection{Downstream Task Evaluation}
\label{main_results}
To evaluate the effectiveness of \our{}, we choose four diverse state of the art algorithms for crystal property prediction, CGCNN~\cite{xie2018crystal}, GATGNN~\cite{louis2020graph}, CrysXPP~\cite{das2022crysxpp} and ALIGNN~\cite{choudhary2021atomistic}. To train these models for any specific property, we adopt the multi-task setting discussed in equation \ref{eq:finetune_loss} ,where we distill relevant knowledge from the pre-trained \our{} to each of these algorithms to predict different properties. 
We report mean absolute error (MAE) of the predicted and actual value of a particular property to compare the performance of different participating methods. For each property, we %split the available dataset into different train,
trained on $80\%$ data, validated on $10\%$ and evaluated on $10\%$ of the data. %validation and test sets and train the property predictor on the training set, tune the hyper-parameters on the validation set and evaluate the performance on the test set. 
We compare the results of distilled version of each SOTA model with its vanilla version (version reported in the respective papers), to show the effectiveness of the proposed framework.\\
\noindent\textbf{Results.} 
\tblak{In Table \ref{tbl-full-data-1}, 
we report MAE of different crystal properties of Materials project and JARVIS-DFT datasets. %, by distilled as well as vanilla version of the SOTA property predictor models, trained on $80\%$ data, validated on $10\%$ and evaluated on $10\%$ of the data. 
In the distilled version of the SOTA models, while training the model, we distill information from the pre-trained \our{} model. We observe that the  distilled version of any state-of-the-art model outperforms the vanilla model across all the properties.  In specific, average improvement in CGCNN, CrysXPP, GATGNN and ALIGNN are 16.20\%, 12.21\%, 8.02\% and 4.19\%, respectively. 
These improvements are particularly significant as in most of the cases, the MAE is already low for SOTA models, still pretraining enables improvement over that.  In fact, lower the MAE, higher the improvement. 
We calculate Spearman's Rank Correlation between MAE for each property across different SOTA models and their improvement due to distilled knowledge and found it to be very high (0.72), which supports  the aforementioned observations.
%These properties depend greatly on crystal structures and atom features, which are being explicitly captured by all the GNN-based SOTA models and our pre-trained \our{} framework. Hence error is lower by the SOTA models and injecting distilled structural information from the pre-trained model is able to achieve much improvements.}\\
% Further we performed detailed analysis of the magnitude of performance improvements and found two different classes of properties. \\
% First, for properties (like formation energy, bandgap, total energy, bulk modulus, etc.), where SOTA models are already performing better (lesser MAE), distilled knowledge from pre-trained model enhances the performance more and achieves a substantial improvement. We report the MAEs for these properties for both the vanilla and distilled model in Table \ref{tbl-full-data-1}. In specific, for these properties, average improvement in CGCNN, CrysXPP, GATGNN and ALIGNN are 23.3\%, 14.8\%, 12.6\% and 7.4\%, respectively. \pg{Put in the same order as in the table.} These properties depend greatly on crystal structures and atom features, which are being explicitly captured by all the GNN-based SOTA models and our pre-trained \our{} framework. Hence error is lower by the SOTA models and injecting distilled structural information from the pre-trained model is able to achieve much improvements.} \\
%Interestingly, we observe that 
The average relative improvement across all properties for ALIGNN (4.19\%) and GATGNN (8.02\%) is lesser compared to CGCNN (16.20\%) and CrysXPP (12.21\%). A possible reason could be that %the design of the encoder for \our{} is simple and similar to CGCNN and CrysXPP. Moreover, 
ALIGNN and GATGNN are more complex models (more number of parameters) than the pre-trained \our{} framework.
Hence designing a deeper pre-training model or additionally incorporating angle-based information (ALIGNN) or attention mechanism (GATGNN) as a part of pre-training framework may help to improve further.  This requires further investigation and we keep it as a scope of future work.}\\
\noindent\textbf{Comparison with Existing Pre-trained Models.} \tblak{We further demonstrate the effectiveness of the knowledge distillation method vis-a-vis the conventional fine-tuning approaches. %Since the parameter space is same for \our{} and CGCNN, we can make two different vesions - (a).
Note that the encoding architecture is same for \our{}, CGCNN, and CrysXPP. CrysXPP is very similar to a {\em pretrained-finetuned} version of CGCNN. 
Thus we compare distilled version of CGCNN  with finetuned version of \our{} and  CrysXPP. 
Additionally, we consider  Pretrain-GNN~\cite{hu2020pretraining} which is a popular pre-training algorithm for molecules. 
% Using techniques specified in CGCNN \cite{xie2018crystal}, we derive a unit molecular structure for a crystal.
%, which are based on pre-training followed by fine-tuning in GNN framework. We consider distilled version of CGCNN for this comparison, because it follows simple graph convolution network on crystal multigraph structure, which is also adopted both in \our{} and CrysXPP that is encoding part is same for all three.
We pre-train all the baseline models on our curated 800K untagged crystal data and fine-tune on seven properties in JARVIS dataset and report the MAE in Table \ref{tbl-distill-finetune}. 
We feed multi-graph structure of the crystal material (as discussed in ``Crystal Graph Representation") in Pretrain-GNN and try different combinations of node-level pre-training strategy along with the graph-level supervised pre-training (as suggested in ~\cite{hu2020pretraining}) and report the minimum MAE for any specific property. For finetuned \our{}, we take the pre-trained encoder of \our{} and feed a multilayer perceptron to predict a specific property. We observe that distilled CGCNN outperforms finetuned version of \our{} and both the baselines with a significant margin for all the properties. Pretrain-GNN performs the worst and the potential reason is - it is designed considering simple two-dimensional structure of molecules with a minimal set of node and bond features, which is hard to generalize for crystal materials which have very complex structure with a rich set of node and edge features.}\\
% \subsection{Effectiveness on sparse training dataset.}
\noindent\textbf{Effectiveness on sparse training dataset.}
Finally, to demonstrate the effectiveness of the pre-training in limited data settings, we conduct additional set of experiments under different training data split. In specific, we vary available training data from 20 to 60 \%, train different SOTA models and check their performance on test dataset. We observe that the distilled version of any SOTA model consistently outperforms its vanilla version even more in the limited training data setting, which illustrates the robustness of our pre-training framework. We report the MAE values of different baselines and their distilled version in Table \ref{tbl-limited-data-small}. 
% Detailed observations are kept in Appendix.
% \noindent\textbf{Analysis of Different Pre-training Losses.}\\
\subsection{Analysis of Different Pre-training Losses}
We perform an ablation study to investigate the influence of different pre-training losses in enhancing the SOTA model performance. While pre-training \our{} (Eq. \ref{eq:pretrain_loss}), we capture both local chemical and global structural information via node and graph-level decoding, respectively. 
Further, we are curious to know the influence of each of these decoding policies independently in the downstream property prediction task.  
\tblak{In specific, we conduct the ablation experiments, where we pre-train \our{} with (a) only node-level decoding ($\mathcal{L}_{FR}$, $\mathcal{L}_{CR}$), (b) only graph-level decoding ($\mathcal{L}_{SG}$, $\mathcal{L}_{NTXent}$). Further, we perform ablations with individual graph-level losses, and pretrain with (c) removing $\mathcal{L}_{NTXent}$ (node-level with $\mathcal{L}_{SG}$ (space group)) and (d) removing $\mathcal{L}_{SG}$ (node-level with $\mathcal{L}_{NTXent}$(crystal system)).} We train two baseline models, CGCNN and ALIGNN, with distilled knowledge from all the aforementioned variants of the pre-trained model and evaluate the performance on four crystal properties. \\
\begin{table}
  \centering
  \small
    \setlength{\tabcolsep}{6 pt}
    \scalebox{0.8}{
      \begin{tabular}{c | c c c c}
        \toprule
        Property & CGCNN & \our{} & CrysXPP & Pretrain \\
        & (Distilled) & (Finetuned) & & -GNN\\
        % & (Distilled) &  &   \\
        \midrule
        Formation Energy  & \textbf{0.047} & 0.056 & 0.062 &  0.764 \\
        Bandgap (OPT)     &  \textbf{0.160} & 0.183 & 0.190 &  0.688 \\
        Total Energy     &  \textbf{0.053} & 0.069 & 0.072 &  1.451 \\
        Ehull             &  \textbf{0.121} & 0.130 & 0.139 &  1.112 \\
        Bandgap (MBJ)     &  \textbf{0.340} & 0.371 & 0.378 &  1.493 \\
        Bulk Modulus (Kv) & \textbf{12.31}  & 13.42 & 13.61 &  20.34 \\
        Shear Modulus (Gv)& \textbf{10.87}  & 11.07 & 11.20 &  16.51 \\
        SLME (\%)         & \textbf{4.791}   & 5.452  & 5.110  &  9.853 \\
        Spillage          & \textbf{0.354}  & 0.374 & 0.363 &  0.481 \\
        \bottomrule
      \end{tabular} 
      }
   \caption{Comparison of the prediction performance (MAE) of seven properties in JARVIS-DFT between \our{} and existing pretrain-finetune models, 
%   All are trained,validated and evaluated on $80-10-10\%$ data respectively.
%   , validated on $10\%$ and evaluated on $10\%$ of the data. 
   the best performance is highlighted in bold.}
  \label{tbl-distill-finetune}
\end{table}
\begin{table}
  \centering
  \small
    \setlength{\tabcolsep}{2.5 pt}
    \scalebox{0.78}{
    \begin{tabular}{c|c|cccc}
    \toprule
    % \multicolumn{2}{c}{Part}                   \\
    % \cmidrule(r){1-2}
    Property & Train-Val & CGCNN & CrysXPP & GATGNN & ALIGNN\\
     & -Test(\%) & (Distilled) & (Distilled) & (Distilled) &(Distilled)\\
    \midrule
     \multirow{3}{*}{\shortstack{Bandgap \\ (MBJ)}}
     & 20-10-70  & 0.453 (23.04)  & 0.450 (24.82) & 0.521 (3.70)  & 0.485 (2.53) \\
     & 40-10-50  & 0.419 (21.41) & 0.405 (18.40) & 0.448 (2.81) & 0.395 (2.20)\\
     & 60-10-30  & 0.364 (19.08) & 0.360 (17.36) & 0.439 (2.29) & 0.380 (1.98)\\
     \midrule
    %  \midrule
     \multirow{3}{*}{\shortstack{Bulk Modulus \\ (Kv)}}
     & 20-10-70  & 16.26 (3.80) & 14.25 (7.59) & 14.19 (4.12)  & 14.06 (4.35)\\
     & 40-10-50  & 14.46 (2.36) & 14.02 (7.34) & 12.59 (3.00)  & 12.11 (2.89)\\
     & 60-10-30  & 14.05 (1.26) & 13.73 (6.98) & 11.75 (2.16) & 11.01 (1.96)\\
     \midrule
    %  \midrule
     \multirow{3}{*}{\shortstack{Shear Modulus \\(Gv)}}
     & 20-10-70  & 12.50 (10.01) & 12.07 (9.86) & 12.42 (3.20)  & 12.31 (3.15)\\
     & 40-10-50  & 11.54 (4.15) & 11.01 (9.46) & 11.23 (1.75)  & 10.67 (2.82)\\
     & 60-10-30  & 11.31 (3.74)  & 10.67 (9.35) & 10.47 (1.69)  & 10.04 (1.95)\\
     \midrule
    %  \midrule
     \multirow{3}{*}{\shortstack{SLME (\%)}}
     & 20-10-70  & 6.62 (7.17) & 5.90 (16.40) & 6.02 (5.20)  & 6.27 (1.43)\\
     & 40-10-50  & 5.78 (5.90) & 5.81 (15.67)  & 5.63 (2.60)  & 5.57 (1.42)\\
     & 60-10-30  & 5.24 (5.68) & 4.84 (10.54)  & 5.34 (2.55)  & 4.82 (1.33)\\
     \midrule
  \end{tabular}
  }
  \caption{MAE values of distilled version of all the SOTA models for four different properties in JARVIS-DFT dataset with the increase in training instances from 20 to 60\%. Relative improvement in the distilled model is mentioned in bracket.}
  \label{tbl-limited-data-small}
\end{table}
\begin{table*}[!thb]
    \centering
    \small
    \setlength{\tabcolsep}{6pt}
    \scalebox{0.75}{
    \begin{tabular}{l c c c c c c c c}
    \toprule
	\textbf{Experiment Settings}  & CGCNN & CGCNN & CrysXPP & CrysXPP & GATGNN & GATGNN & ALIGNN & ALIGNN\\
	 &  & (Distilled) &  & (Distilled) &  & (Distilled) &  & (Distilled)\\
	\midrule
	\textbf{\vtop{\hbox{\strut Train on DFT }\hbox{\strut Test on Experimental}}} 
    & 0.265 & 0.244 (7.60) & 0.243 & 0.225 (7.40) & 0.274 & 0.232 (15.3) 	& 0.220 & 0.209 (5.05)  \\
	\midrule
	\textbf{\vtop{\hbox{\strut Train on DFT and 20 \% Experimental }\hbox{\strut Test on 80 \% Experimental}}}  
	 & 0.144 & 0.113 (21.7) & 0.138 & 0.118 (14.2) & 0.173 &  0.168 (2.70) & 0.099 & 0.094 (5.60) \\
	\midrule
	\textbf{\vtop{\hbox{\strut Train on DFT and 80 \% Experimental }\hbox{\strut Test on 20 \% Experimental}}} 
	 & 0.094 & 0.073 (22.7) & 0.087 & 0.071 (18.4) & 0.113  & 0.109 (3.40) & 0.073 & 0.069 (5.90) \\
    \bottomrule
    \end{tabular}
	}
	\caption{MAE of predicting experimental values by different SOTA models and their distilled versions with full DFT data and  different percentages of experimental data for formation energy in OQMD-EXP dataset. Relative improvement in the distilled model is mentioned in bracket.}
    \label{tab:fe}
\end{table*}
\begin{figure}[!thb]
	\centering
	\boxed{\includegraphics[width=\columnwidth]{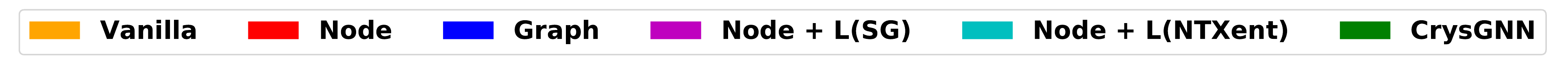}}
	\\
	\vspace*{-1mm}
	\subfloat[Formation Energy]{
		\boxed{\includegraphics[width=0.44\columnwidth, height=24mm]{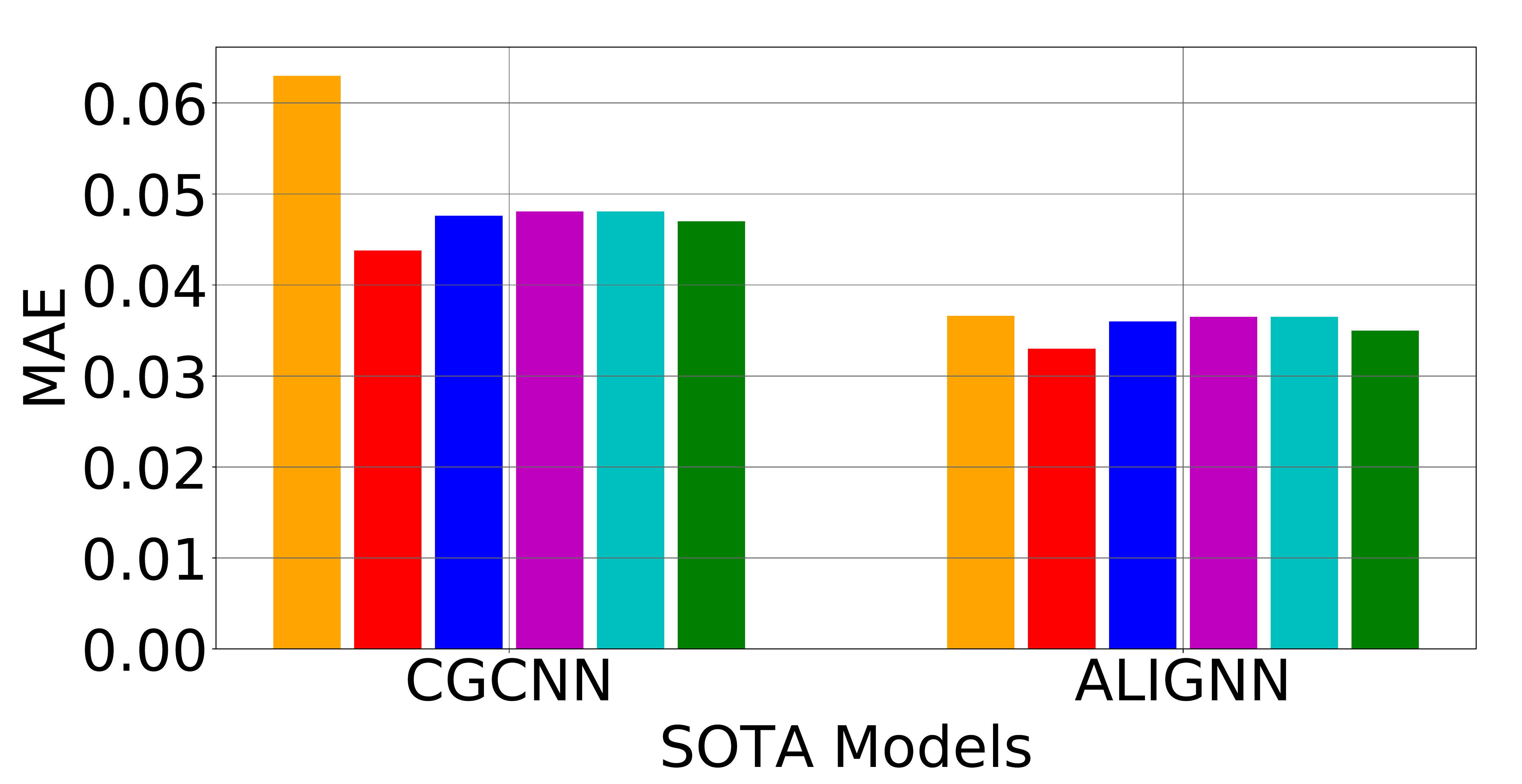}}}
	\subfloat[Bandgap (MBJ)]{
		\boxed{\includegraphics[width=0.44\columnwidth, 
		height=24mm]{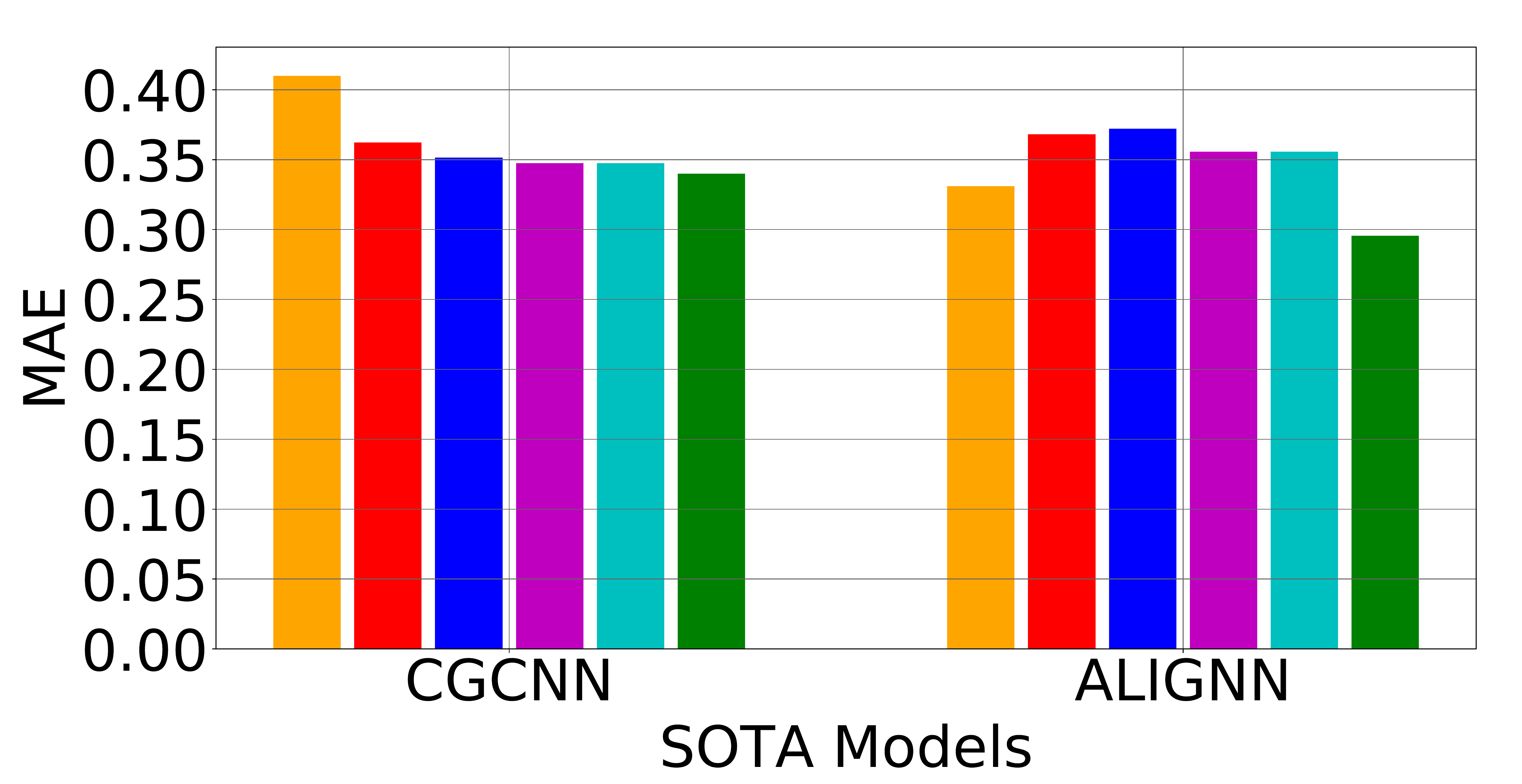}}}\\
	\subfloat[Total Energy]{
		\boxed{\includegraphics[width=0.44\columnwidth, height=24mm]{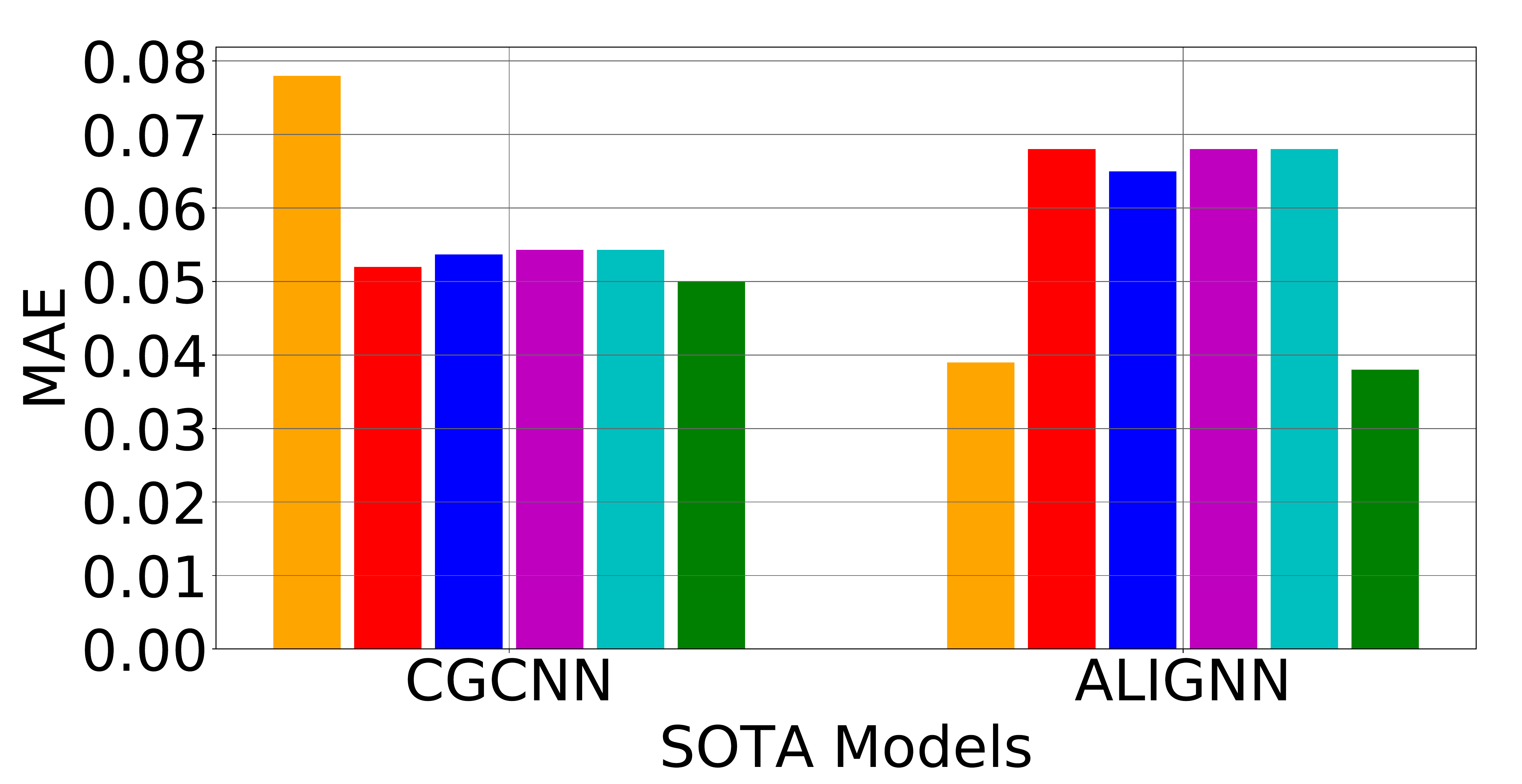}}}
	\subfloat[Bulk Modulus (Kv)]{
		\boxed{\includegraphics[width=0.44\columnwidth, height=24mm]{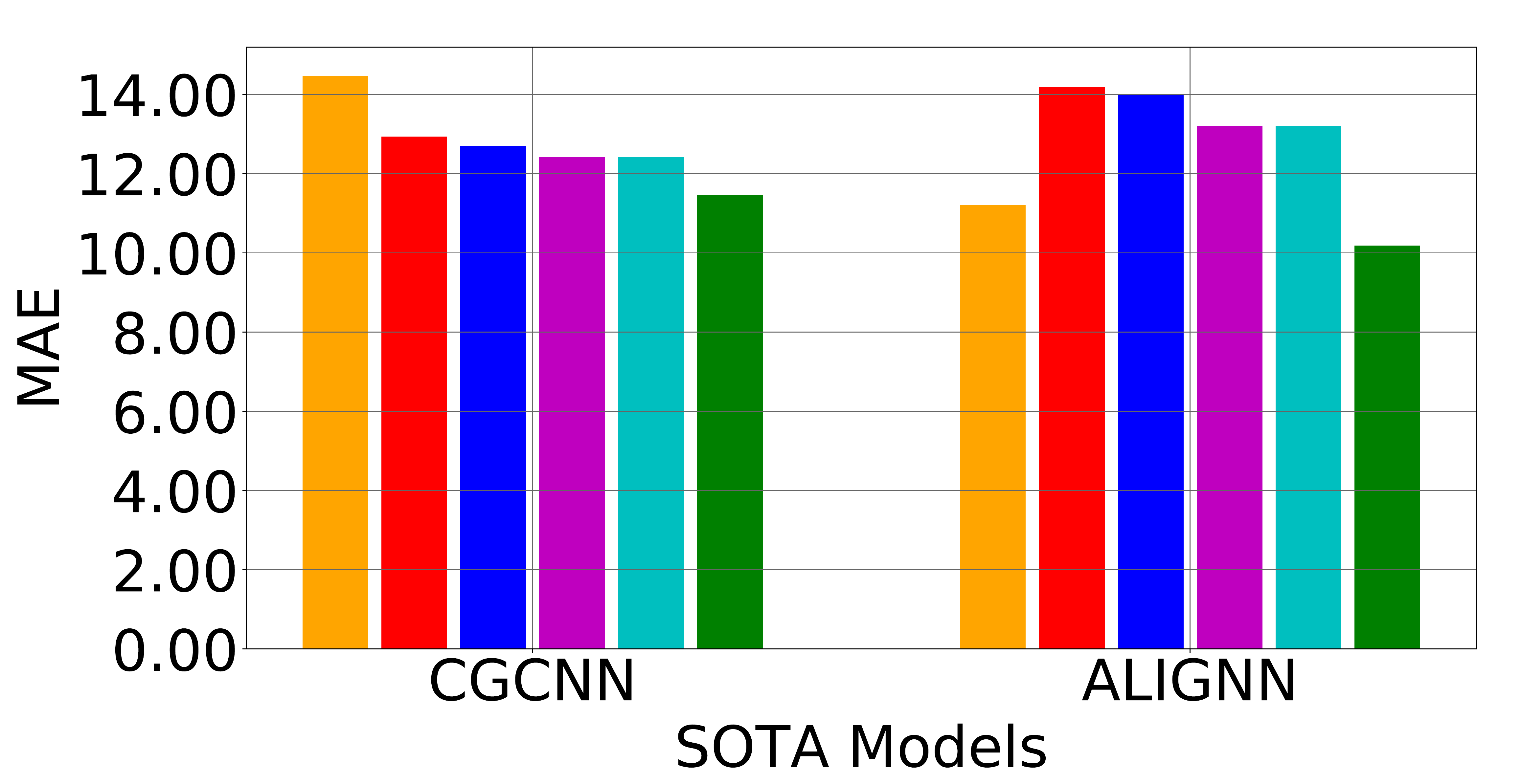}}}
	\caption{\tblak{Summary of experiments of ablation study on importance of different pre-training loss components on \our{} training and eventually its effect on CGCNN and ALIGNN models on four different properties (MAE for property prediction). (i) Vanilla: SOTA based model (without distillation) and all the other cases are SOTA models (distilled) from different pre-trained version of \our{}.
	(ii) Node : only node-level pre-training, (iii) Graph : only graph-level pre-training, (iv) Node + L(SG) : node-level and $\mathcal{L}_{SG}$, (v) Node + L(NTXent) : node-level and $\mathcal{L}_{NTXent}$ and (vi) \our{}~: both node and graph-level pre-training.}}
	\label{fig:ablation_diff_loss}
\end{figure}
Experimental results are presented in Fig.~\ref{fig:ablation_diff_loss}. %\pg{Change both to CrysGNN in figure}
We can observe clearly that all the variants offer significant performance gain in all four properties using the combined node and graph-level pre-training, compared to node-level or graph-level pre-training separately. Only exception is formation energy, where only node-level pre-training produces less error compared to other variants, in both the baseline. Formation energy of a crystal is defined as the difference between the energy of a unit cell comprised of $N$ chemical species and the sum of the chemical potentials of all the $N$ chemical species. Hence pre-training at the node-level (node features and connection) is adequate for enhancing performance of formation energy prediction and incorporating graph-level information works as a noisy information, which degrades the performance. \tblak{We also observe improvement in performance using both supervised and contrastive graph-level losses ($\mathcal{L}_{SG}$ and $\mathcal{L}_{NTXent}$), compared to using only one of them, which proves the learned representation via supervised and contrastive learning is more expressive that using any one of them.}
Moreover, in ALIGNN, with either node or graph-level pre-training separately, performance degrades across different properties. ALIGNN explicitly captures the three body interactions which drive its performance, to replicate that inclusion of both node and graph information is necessary.
% \noindent\textbf{Removal of DFT error bias using experimental data}\\
\subsection{Removal of DFT error bias using experimental data}
One of the fundamental issues in material science is that experimental data for crystal properties are very rare~\cite{kubaschewski1993materials,bracht1995properties,turns1995understanding}. Hence existing SOTA models rely on DFT calculated data to train their parameters. % for better property prediction accuracy. 
However, mathematical approximations in DFT calculation lead to erroneous predictions (error bias) compared to the actual experimental values of a particular property. Hence %current SOTA models can not be extended to predict as accurate as the experimental results and 
DFT error bias is prevalent in all SOTA models. %\tblu{However, it is empirically known 
\cite{das2022crysxpp} has shown that pre-training helps to remove DFT error bias when fine-tuned with experimental data. Hence, %we are curious to understand whether distilling prior knowledge from the pre-trained \our{} could aid SOTA models to mitigate this issue. \\
%In this section, we will we 
we investigate whether SOTA models can remove the DFT error with distilled knowledge from pre-trained model, using a small amount of available experimental data. In specific, we consider OQMD-EXP dataset to conduct an experiment, where we train SOTA models and their distilled variants with available DFT data and different percentages of experimental data for formation energy. We report the MAE of different SOTA models and their distilled variant in Table \ref{tab:fe}.  We observe, with more amount of experimental training data, all the SOTA models are minimizing the error consistently. Moreover, with distilled knowledge from pre-trained \our{}, all SOTA models are reducing MAE further and we observe consistently larger degree of improvement with more amount of experimental training data in almost all the models. % (except GATGNN).

\section{Conclusion}
\label{conclusion}
In this work, we present a novel but simple pre-trained GNN framework, \our{}, for crystalline materials, which captures both local chemical and global structural semantics of crystal graphs. To pre-train the model, we curate a huge dataset of  800k unlabelled crystal graphs. %and adopted both node-level and graph-level pre-training strategies to learn more expressive node and graph-level embeddings. 
Further, while predicting different crystal properties, we distill important knowledge from \our{} and inject it into different state of the art property predictors and enhance their performance. Extensive experiments on multiple popular datasets and diverse set of SOTA models show that with distilled knowledge from the pre-trained model, all the SOTA models outperform their vanilla versions.
%on 19 different properties. 
% We further observe that distilled pre-trained knowledge improves the performance of SOTA models even better with limited tagged data. 
Extensive experiments show its superiority over conventional fine-tune models and its inherent ability to remove DFT-induced bias.
The pretraining framework can be extended beyond structural graph information in a multi-modal setting to include other important (text and image) information about a crystal which would be our immediate future work.
%tWe further observe that the distillation process provides significant improvement over the conventional approach of finetuning the pre-trained model.
%Finally, we observe that the proposed framework can heavily mitigate the DFT bias error in the state of the art models using a small amount of experimental data.\\

%\noindent{\bf Limitations and Future Work.}
%Though our proposed pre-trained model is able to enhance the performance of all the state of the art baseline models, improvements for all types of SOTA models are not consistent and we found lesser improvement in case of complex models like GATGNN and ALIGNN. This provides scope for further investigation on designing a more complex and deeper pre-trained model. We need to explore different graph representations, loss variations, distillation strategies, etc. Of course, the very fact that the simple model, \our{}, can provide substantial improvements, lays the foundation  for such future exploration.  Moreover, in this present work, we have focused on predicting crystal properties, which is a graph-level regression task. However, the same framework can be used to  explore the effect of pre-training on other graph-level tasks (classification or clustering) or node-level tasks which would be one of our future work.  

\section{Acknowledgments}
This work was partially funded by Indo Korea Science and Technology Center, Bangalore, India, under the project name {\em Transfer learning and Weak Supervision for Accurate and Interpretable Prediction of Properties of Materials from their Crystal Graph Representation}. Also, it was (partially) funded by the Federal Ministry of Education and Research (BMBF), Germany under the project LeibnizKILabor with grant No.\,01DD20003. We thank the Ministry of Education, Govt of India, for supporting Kishalay with Prime Minister Research Fellowship during his Ph.D. tenure.
\bibliography{aaai23}
\clearpage
% CrysGNN : Distilling pre-trained knowledge to enhance property prediction for crystalline materials.(Appendix)
\section{Related Work}
\subsection{Materials Property Prediction.}
In recent times, data-driven approaches have become quite popular to establish relationships between the atomic structure of crystalline materials and their properties with very high precision. Especially, graph neural network (GNN) based approaches have emerged as a powerful machine learning model tool to encode material’s complex topological structure along with  node features in an enriched representation space. Models such as CGCNN~\cite{xie2018crystal} represent 3D crystal structure as a multi-graph and build a graph convolution neural network directly on the graph to update node features based on their local chemical and structural environment. GATGNN~\cite{louis2020graph} incorporates the idea of graph attention network to learn the importance of different inter atomic bonds whereas MEGNet~\cite{chen2019graph} introduces global state attributes for quantitative structure-state property relationship prediction in materials. While all these models implicitly represent many body interactions through multiple graph convolution layers, ALIGNN~\cite{choudhary2021atomistic} explicitly captures many body interactions by incorporating bond angles and local geometric distortions.
\subsection{Graph Pre-training Strategies}
There are attempts to pre-train GNNs to extract graph and node level representations. ~\cite{hu2020pretraining} develops an effective pre-training strategy for GNNs, where they perform both node level and graph level pre-training on GNNs to capture domain specific knowledge about nodes and edges, in addition to global graph-level knowledge. GPT-GNN~\cite{hu2020gpt} proposes a self-supervised attributed graph generation task to pre-train a GNN model, which captures structural and semantic properties of the graph. ~\cite{qiu2020gcc} presents GCC, which leverages the idea of contrastive learning to design the graph pre-training task as subgraph instance discrimination, to capture universal network topological properties across multiple networks. Another recent work is GraphCL~\cite{you2020graph}, which proposes different graph augmentation methods and maximises the agreement between two augmented views of the same graph via a contrastive loss. 
In the field of crystal graphs, CrysXPP~\cite{das2022crysxpp} is the only model which comes close to a pre-trained model. In their work, an autoencoder is trained on  a  volume of un-tagged crystal graphs and  the learned knowledge is (transferred to)  used to initialize the encoder of  CrysXPP, which is fine-tuned with property specific tagged data. \\
To the best of our knowledge, \our{} is the first attempt to develop a deep pre-trained model for the crystalline materials and for that, we curated a new large untagged crystal dataset with 800K crystal graphs.
\section{Crystal Representation}
We realize a crystal material as a multi-graph structure $\mathcal{G}_i =(\mathcal{V}_i, \mathcal{E}_i, \mathcal{X}_i, \mathcal{F}_i )$ as proposed in \cite{das2022crysxpp,xie2018crystal}. $\mathcal{G}_i$ is an undirected weighted multi-graph where $\mathcal{V}_i$ denotes the set of nodes or atoms present in a unit cell of the crystal structure. $\mathcal{E}_i=\{(u,v,k_{uv})\}$ denotes a multi-set of node pairs and $k_{uv}$ denotes number of edges between a node pair $(u,v)$. $\mathcal{X}_i=\{(x_{u} | u \in \mathcal{V}_i )\}$ denotes 92 dimensional node feature set proposed by CGCNN~\cite{xie2018crystal}. It includes different chemical properties like electronegativity, valance electron, covalent radius, etc. Details of these node features are given in Table \ref{tab:feature_desc}. Finally, $\mathcal{F}_i=\{\{s^k\}_{(u,v)} |  (u,v) \in \mathcal{E}_i, k\in\{1..k_{uv}\}\}$ denotes the multi-set of edge weights where $s^k$ corresponds to the $k^{th}$ bond length between a node pair $(u,v)$, which signifies the inter-atomic bond distance between two atoms.
\begin{figure}[h]
	\centering
	\vspace*{-1mm}
	\boxed{\includegraphics[width=\columnwidth, height=45mm]{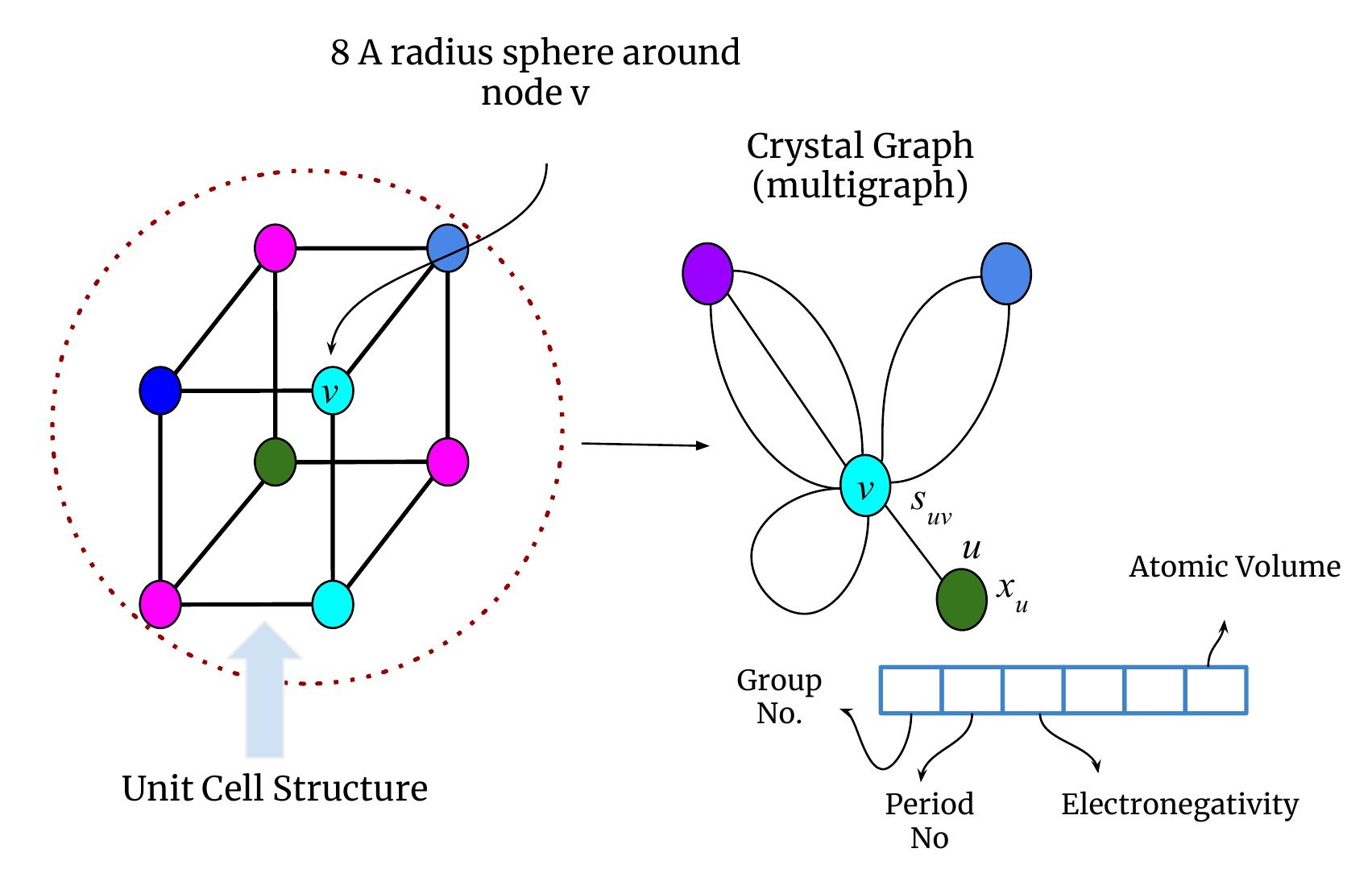}}
	\caption{Multi-graph Representation from Crystal 3D Structure.}
	\label{fig:3d_crsytal_rep}
\end{figure}
\begin{table}[]
	\centering
	\small
	\setlength{\tabcolsep}{4pt}
	\scalebox{0.9}{
	\begin{tabular}{|c|c|c|c|}
		\hline
		\textbf{Features} & \textbf{Unit} &\textbf{Range}  & \textbf{Feature Dimension} \\
		\hline
		Group Number & - & 1,2, ..., 18 & 18  \\
		\hline
		Period Number & - & 1,2, ..., 9 & 9\\
		\hline
		Electronegativity & - & 0.5-4.0 & 10 \\
		\hline
		Covalent Radius & pm & 25-250 & 10 \\
		\hline
		Valence Electrons & - &1,2, ..., 12 & 12\\
		\hline
		First Ionization Energy & eV & 1.3–3.3 & 10\\
		\hline
		Electron Affinity & eV & -3–3.7 & 10 \\
		\hline
		Block & - & s, p, d, f & 4 \\
		\hline
		Atomic Volume & $cm^{3} /mol$ & 1.5–4.3 & 10 \\
		\hline
	\end{tabular}
	}
	\caption{Description of different atomic features $\mathcal{X}_i$ and their unit, range and dimensions. All of them together forms 92 dimensional node feature vector $\mathcal{X}_i$.}
    \centering
	\label{tab:feature_desc}
\end{table}
\section {Details of GNN Architecture in \our{}}
Here we describe GNN architectures used in CrysGNN for pre-training and property prediction model.
\subsubsection{Encoder.} We develop an graph convolution \cite{xie2018crystal} based encoding module, which takes crystal multi-graph structure $\mathcal{G} =(\mathcal{V}, \mathcal{E}, \mathcal{X}, \mathcal{F} )$ as input and encode structural semantics of the crystal graph into lower dimensional space. Considering $L$ graph convolution layers, neighbourhood aggregation of the $l$-th layer is represented as :
\begin{equation}
    \label{eq:cgcnn}
    z_u^{(l+1)}  = z_u^{(l)} + \sum\limits_{v,k} \sigma ( {h^{(l)}_{(u,v)}}_k W_c^{(l)} +  b_c^{(l)}) \odot g({h^{(l)}_{(u,v)}}_k W_s^{(l)}+b_s^{(l)}) 
\end{equation}
where $h_{(u,v)_k}^{(l)} = z_u^{(l)} \oplus z_v^{(l)} \oplus {s^{(k)}_{(u,v)}}$. Here $s^{(k)}_{(u,v)}$ represents inter-atomic bond length of the $k$-th bond between $(u,v)$ and $z_u^{(l)}$ denotes embedding of node $u$ after $l$-th layer, which is initialized to a transformed node feature vector $z_u^0 := x_u  W_x$ where $W_x$ is the trainable parameter and $ x_u $ is the input node feature vector (Table. \ref{tab:feature_desc}). In Equation \ref{eq:cgcnn}, $W_c^{(l)},W_s^{(l)},b_c^{(l)},b_s^{(l)}$ are the convolution weight matrix, self weight matrix, convolution bias, and self bias of the $l$-th layer, respectively, $\oplus$ operator denotes concatenation, $\odot$ denotes element-wise multiplication, $\sigma$ is the sigmoid function indicating the edge importance and $g$ is a feed forward network. After $L$-layers of such propagation, $\mathcal{Z} = \{z_1,..., z_{|\mathcal{V}|}\}$ represents the set of final node embeddings, where $z_u :=z_u^L$ represents final embedding of node $u$.
\subsubsection{Decoder.} In the decoding module, we apply both node-level and graph-level self-supervised pre-training as discussed in \textbf{Node Level Decoding} and \textbf{Graph Level Decoding} sections in the main manuscript.
\subsubsection{Property Predictor.} For the property prediction model, we use four diverse sets of SOTA models, which have different encoder choices and on top of that they have a regressor module to predict the property value. We retrofit \our{} model with each of these SOTA models using the modified multitask loss (incorporating knowledge distillation loss with their property loss) as discussed in Equation \ref{eq:finetune_loss} (main manuscript).
\section{Crystal System and Space group Information}
By definition crystal is a periodic arrangement of repeating “motifs”( e.g. atoms, ions). The symmetry of a periodic pattern of repeated motifs is the total set of symmetry operations allowed by that pattern. The total set of such symmetry operations, applicable to the pattern is the pattern's symmetry, and is mathematically described by a so-called Space Group. The Space Group of a Crystal describes the symmetry of that crystal,
and as such it describes an important aspect of that crystal's internal structure.\\
In turn  crystals across certain space groups show
similarities among each other, hence they are divided among seven Crystal Systems : Triclinic, Monoclinic, Orthorhombic, Tetragonal, Trigonal, Hexagonal, and Cubic or Isometric system ~\cite{kittel2018kittel}.\\
\textbf{Crystallographic Axes: } The crystallographic axes are imaginary lines that we can draw within the crystal lattice.  These will define a coordinate system within the crystal. We refer to the axes as \textbf{a}, \textbf{b}, \textbf{c} and angle between them as $\alpha$, $\beta$ and $\gamma$. By using these crystallographic axes we can define six large groups or crystal systems.
% \noteng{Please give reference from where you have copied}
\begin{itemize}
  \item \textbf{Cubic or Isometric system :} The three crystallographic axes are all equal in length and intersect at right angles to each other.
  \begin{equation}
        \boxed{a=b=c\;; \;\; \alpha = \beta =\gamma = 90 \degree} \notag
    \end{equation}
  \item \textbf{Tetragonal system :} Three axes, all at right angles, two of which are equal in length (a and b)
    and one (c) which is different in length (shorter or longer).
    \begin{equation}
        \boxed{a=b\neq c\;; \;\; \alpha = \beta =\gamma = 90 \degree} \notag
    \end{equation}
   \item \textbf{Orthorhombic system :} Three axes, all at right angles, all three have different length.
   \begin{equation}
        \boxed{a\neq b\neq c\;; \;\; \alpha = \beta =\gamma = 90 \degree} \notag
    \end{equation}
   \item \textbf{Monoclinic system :} Three axes, all unequal in length, two of which (a and c) intersect at an oblique angle (not 90 degrees), the third axis (b) is perpendicular to the other two axes.
   \begin{equation}
        \boxed{a\neq b\neq c\;; \;\; \alpha  =\gamma = 90 \degree \neq \beta} \notag
    \end{equation}
   \item \textbf{Triclinic system :} The three axes are all unequal in length and intersect at three different angles (any angle but 90 degrees).
   \begin{equation}
        \boxed{a\neq b\neq c\;; \;\; \alpha \neq \beta \neq \gamma \neq 90 \degree} \notag
    \end{equation}
  \item \textbf{Hexagonal system :} Here we have four axes. Three of the axes fall in the same plane and intersect at the axial cross at 120°. These 3 axes, labeled $a_1$, $a_2$, and $a_3$, are the same length. The fourth axis, c, may be of {\em a} different length than the axes set. The c axis also passes through the intersection of the {\em a} axes
set at right angle to the plane formed by the {\em a} set.
\end{itemize}

\section{Details about datasets and different crystal properties.}
\textbf{OQMD:} The Open Quantum Materials Database (OQMD) is a high-throughput database based on DFT calculations of chemicals from the Inorganic Crystal Structure Database (ICSD) and embellishments of regularly occurring crystal structures \cite{OQMD}. Comprehensive computations for significant structure types in materials research, such as Heusler and perovskite compounds, are among the distinguishing aspects of this database. The database is public and  a free resource.\\\\
\textbf{Materials project:} This is another free and public resource database that contains crystal structures and calculated materials properties~\cite{MP}. The dataset is made from the results obtained with density functional theory based calculations. The dataset is composed of electronic structure, thermodynamic, mechanical, and dielectric properties. It also provides  a visual web-based search interface.\\\\
\textbf{JARVIS:} JARVIS (Joint Automated Repository for Various Integrated Simulations) is a data repository that incorporates not only DFT-based calculations, but also data from classical force-fields and machine learning approaches~\cite{JARVIS}. The databases are free and public as well.\\\\
We curated around {800K untagged crystal graph data from two popular materials databases, Materials Project (MP) and OQMD}, to pre-train \our{} model. Further to evaluate the performance of different SOTA models with distilled knowledge from \our{}, we select {MP 2018.6.1 version of Materials Project} and  {2021.8.18 version of JARVIS-DFT}, another popular materials database, for property prediction as suggested by \cite{choudhary2021atomistic}.  Please note, MP 2018.6.1 dataset is a subset of the dataset used for pre-training, whereas JARVIS-DFT is a separate dataset which is not seen during the pre-training.
MP 2018.6.1 consists of 69,239 materials with two properties namely bandgap and formation energy, whereas JARVIS-DFT consists of 55,722 materials with 19  properties which can be broadly classified into two categories : 1) properties like formation energy, bandgap, total energy, bulk modulus, etc. which depend greatly on crystal structures and atom features, and 2) properties like $\epsilon_x$, $\epsilon_y$, $\epsilon_z$, n-Seebeck, n-PF, etc. which depend on the precise description of the materials’ electronic structure. Details of these properties is provided in Table \ref{tbl-prop} \\
Moreover, all these properties in both Materials Project and JARVIS-DFT datasets are based on DFT calculations of chemicals. Therefore, to investigate how pre-trained knowledge helps to mitigate the DFT error, we also take a small dataset ~\cite[OQMD-EXP]{kirklin2015open} containing 1,500 available experimental data of formation energy. Details of each of these datasets are given in Table \ref{tbl-dataset}.
\begin{table}
  \centering
  \small
    \setlength{\tabcolsep}{10 pt}
  \begin{tabular}{ccc}
    \toprule
    % \multicolumn{2}{c}{Part}                   \\
    % \cmidrule(r){1-2}
    Property & Unit & Data-size\\
    \midrule
     Formation\_Energy & $eV/(atom)$ &  69239 \\
     Bandgap (OPT)    & $eV$ & 69239  \\
    \midrule
    Formation\_Energy & $eV/(atom)$ & 55723 \\
    Bandgap (OPT)    & $eV$ & 55723 \\
    Total\_Energy    & $eV/(atom)$ & 55723\\
    Ehull    & $eV$ & 55371\\
    Bandgap (MBJ)    & $eV$ & 18172 \\
    
    Bulk Modulus (Kv)    & GPa & 19680\\
    Shear Modulus (Gv)    & GPa & 19680\\
    SLME (\%)    & No unit & 9068\\
    Spillage    & No unit & 11377 \\
    $\epsilon_x$ (OPT)    & No unit & 44491 \\
    $\epsilon_y$ (OPT)    & No unit & 44491\\
    $\epsilon_z$ (OPT)    & No unit & 44491 \\
    $\epsilon_x$ (MBJ)    & No unit & 16814\\
    $\epsilon_y$ (MBJ)    & No unit & 16814\\
    $\epsilon_z$ (MBJ)    & No unit & 16814 \\
    n-Seebeck & $\mu VK^{-1}$ & 23210\\
    n-PF & $\mu W(mK^2)^{-1}$ & 23210\\
   p-Seebeck & $\mu VK^{-1}$ &  23210\\
    p-PF & $\mu W(mK^2)^{-1}$ & 23210\\
     \bottomrule
  \end{tabular}
  \caption{Summary of different crystal properties in Materials Project (Top)  and JARVIS-DFT (Bottom)  datasets.}
   \label{tbl-prop}
\end{table}
\begin{table*}
  \centering
  \small
    \setlength{\tabcolsep}{11 pt}
    \scalebox{0.9}{
      \begin{tabular}{c | c c | c c| c c | c c}
        \toprule
        Property  & CGCNN & CGCNN & CrysXPP & CrysXPP & GATGNN & GATGNN & ALIGNN & ALIGNN\\
        &  & (Distilled) &  & (Distilled) &  & (Distilled) &  & (Distilled)\\
        \midrule
        $\epsilon_x$ (OPT)    & 24.93 & \textbf{23.86} & 24.47 & \textbf{23.33} & 26.79 & \textbf{25.23}  & 21.42 & \textbf{21.37}\\
        $\epsilon_y$ (OPT)    & 25.06 & \textbf{24.08} & 24.84 & \textbf{23.98} & 26.77 & \textbf{25.02}  & 21.66 & \textbf{21.25}   \\
        $\epsilon_z$ (OPT)    & 24.99 & \textbf{23.85} & 24.13 & \textbf{23.76} & 26.09 & \textbf{24.81}  & \textbf{20.51} & 21.03 \\
        $\epsilon_x$ (MBJ)     & 27.18 & \textbf{26.64} & 28.40 & \textbf{26.03} & \textbf{27.56} & 27.61 & 24.76 & \textbf{24.32}\\
        $\epsilon_y$ (MBJ)     & 27.14 & \textbf{25.80} & 26.92 & \textbf{25.34} & 27.21 & \textbf{26.89}  & 23.99 & \textbf{23.83}\\
        $\epsilon_z$ (MBJ)     & 28.38 & \textbf{26.08} & 26.76 & \textbf{25.97} & 25.50 & \textbf{24.84}  & \textbf{23.93} & 24.59\\
        n-Seebeck  & 50.32 & \textbf{46.66} & 48.79 & \textbf{46.53} & 52.03 & \textbf{49.46}  & 42.85 & \textbf{42.13}\\
        n-PF  & 528.25 & \textbf{506.41} & 502.79 & \textbf{498.43} & 511.56 & \textbf{505.22} & 477.09 & \textbf{474.88} \\
      p-Seebeck  & 53.01 & \textbf{49.15} & 50.13 & \textbf{48.67} & 54.99 & \textbf{50.72}  & 46.04 & \textbf{45.19}\\
        p-PF  & 513.57 & \textbf{501.11} & 493.35 & \textbf{489.54} & 516.84 & \textbf{494.92}  & 460.11 & \textbf{447.81}\\
         \bottomrule
      \end{tabular}
      }
  \caption{Summary of the prediction performance (MAE) of different properties (belong to second class) in JARVIS-DFT dataset where performance improvement is modest. Model M is the vanilla variant of a SOTA model and M  (Distilled) is the distilled variant using the pretrained \our{}. All the models are trained on $80\%$  data, validated on $10\%$ and evaluated on $10\%$ of the data. The best performance is highlighted in bold.}
  \label{tbl-full-data-2}
\end{table*}
% \section{Training Setup / Hyper-parameter Details.}
\section{Training Setup / Hyper-parameter Details / Computational Resources used.}
We use five convolution layers of the encoder module to train \our{} and train it for 200 epochs using Adam \cite{kingma2014adam} for optimization with a learning rate of 0.03. We keep the embedding dimension for each node as 64, batch size of data as 128, and equal weightage (0.25) for $\alpha$, $\beta$, $\gamma$, and $\lambda$ of Equation \ref{eq:pretrain_loss}. During property prediction, we used the default configuration for hyper-parameters for all four vanilla SOTA models. However, while predicting properties by the distilled version of the SOTA models we keep node embedding dimension for each SOTA as 64, and train each model for 500 epochs using Adam~\cite{kingma2014adam} optimization and keep $\delta$ as 0.5. We perform the experiments in shared servers having Intel E5-2620v4 processors which contain 16 cores/thread  and four GTX 1080Ti 11GB GPUs each.
\section{Baseline Models.}
To evaluate the effectiveness of \our{}, we choose following four diverse state of the art algorithms for crystal property prediction.
\begin{enumerate}
	\item \textbf{CGCNN}~\cite{xie2018crystal} : This work generates crystal graphs from inorganic crystal materials and builds a graph convolution based supervised model for predicting various properties of the crystals.
	\item \textbf{GATGNN}~\cite{louis2020graph} : In this work, authors have incorporated a graph neural network with multiple graph-attention layers (GAT) and a global attention layer, which can learn efficiently the importance of different complex bonds shared among the atoms within each atom’s local neighborhood.
	\item \textbf{ALIGNN}~\cite{choudhary2021atomistic} : In this work, authors adopted line graph neural networks to develop an alternative way to
    include angular information into convolution layer which alternates between message passing on the bond graph and its bond-angle line graph.
	\item \textbf{CrysXPP}~\cite{das2022crysxpp} : In this work,  the authors train an autoencoder (CrysAE) on  a  volume of un-tagged crystal graphs and then the learned knowledge is (transferred to)  used to initialize the encoder of  CrysXPP, which is fine-tuned with property specific tagged data.Also they design a feature selector that helps to interpret the model’s prediction. 
\end{enumerate}
\subsection{Baseline Implementations.} We used the available PyTorch implementations of all the baselines, viz, CGCNN\footnote{https://github.com/txie-93/cgcnn.git}, GATGNN\footnote{https://github.com/superlouis/GATGNN.git}, CrysXPP\footnote{https://github.com/kdmsit/crysxpp.git} and ALIGNN\footnote{https://github.com/usnistgov/alignn.git}. In order to retrofit \our{} to each SOTA model, we modified their supervised loss into the multi-task loss proposed in Equation \ref{eq:finetune_loss} (main manuscript) and train each model. To ensure a fair comparison between all the baselines and perform the knowledge distillation between node embeddings, we have ensured node embedding dimension for each SOTA and \our{} as 64.
\begin{table*}
  \centering
  \small
    \setlength{\tabcolsep}{6 pt}
    \scalebox{0.9}{
    \begin{tabular}{cccccccccc}
    \toprule
    % \multicolumn{2}{c}{Part}                   \\
    % \cmidrule(r){1-2}
    Property & Train-Val  & CGCNN & CGCNN & CrysXPP & CrysXPP & GATGNN & GATGNN & ALIGNN & ALIGNN\\
     & -Test(\%) &  & (Distilled) &  & (Distilled) &  & (Distilled) &  & (Distilled)\\
    \midrule
     \multirow{3}{*}{\shortstack{Bandgap \\ (MBJ)}}
     & 20-10-70  & 0.588 & 0.453\textbf{*} (23.04) & 0.598 & 0.450\textbf{*} (24.82) & 0.541 & 0.521 (3.70) & 0.497 & 0.485 (2.53) \\
     & 40-10-50  & 0.532 & 0.419\textbf{*} (21.41) & 0.496 & 0.405\textbf{*} (18.40) & 0.462 & 0.448\textbf{*} (2.81) & 0.404 & 0.395 (2.20)\\
     & 60-10-30  & 0.449 & 0.364 (19.08) & 0.435 & 0.360 (17.36) & 0.449 & 0.439 (2.29) & 0.387 & 0.380 (1.98)\\
     \midrule
    %  \midrule
     \multirow{3}{*}{\shortstack{Bulk Modulus \\ (Kv)}}
     & 20-10-70  & 16.91 & 16.26 (3.80) & 15.42 & 14.25\textbf{*} (7.59) & 14.80 & 14.19 (4.12) & 14.70 & 14.06 (4.35)\\
     & 40-10-50  & 14.81 & 14.46 (2.36) & 15.13 & 14.02\textbf{*} (7.34) & 12.98 & 12.59 (3.00) & 12.47 & 12.11 (2.89)\\
     & 60-10-30  & 14.23 & 14.05 (1.26) & 14.76 & 13.73 (6.98) & 12.01 & 11.75 (2.16) & 11.23 & 11.01 (1.96)\\
     \midrule
    %  \midrule
     \multirow{3}{*}{\shortstack{Shear Modulus \\(Gv)}}
     & 20-10-70  & 13.89 & 12.50 (10.01) & 13.39 & 12.07\textbf{*} (9.86) & 12.83 & 12.42 (3.20) & 12.71 & 12.31 (3.15)\\
     & 40-10-50  & 12.04 & 11.54\textbf{*} (4.15) & 12.16 & 11.01\textbf{*} (9.46)  & 11.43 & 11.23 (1.75) & 10.98 & 10.67 (2.82)\\
     & 60-10-30  & 11.75 & 11.31 (3.74) & 11.77 & 10.67 (9.35) & 10.65 & 10.47 (1.69) & 10.24 & 10.04 (1.95)\\
     \midrule
    %  \midrule
     \multirow{3}{*}{\shortstack{SLME (\%)}}
     & 20-10-70  & 7.13 & 6.62 (7.17) & 7.05 & 5.90\textbf{*} (16.40) & 6.35 & 6.02 (5.20) & 6.36 & 6.27 (1.43)\\
     & 40-10-50  & 6.14 & 5.78 (5.90) & 6.89 & 5.81 (15.67) & 5.78 & 5.63 (2.60) & 5.65 & 5.57 (1.42)\\
     & 60-10-30  & 5.55 & 5.24 (5.68) & 5.41 & 4.84 (10.54) & 5.48 & 5.34 (2.55) & 4.88 & 4.82 (1.33)\\
     \midrule
    %  \midrule
     \multirow{3}{*}{\shortstack{Spillage}}
     & 20-10-70  & 0.424 & 0.389\textbf{*} (8.43) & 0.422 & 0.403 (4.45) & 0.406 & 0.392\textbf{*} (3.5) & 0.402 & 0.392 (2.49)\\
     & 40-10-50  & 0.408 & 0.391\textbf{*} (4.28) & 0.398 & 0.384 (3.57) & 0.396 & 0.388 (1.94) & 0.378 & 0.372 (1.72)\\
     & 60-10-30  & 0.395 & 0.380 (3.80) & 0.379 & 0.368 (3.00) & 0.382 & 0.377 (1.36) & 0.362 & 0.357 (1.13)\\
     
    \bottomrule
  \end{tabular}
  }
  \caption{MAE values of five different properties in JARVIS-DFT dataset with the increase in training instances from 20 to 60\%. Model M is the vanilla variant of the SOTA model, M (Distilled) is distilled variant of it and relative improvement is mentioned in bracket. Distilled knowledge from \our{} improves the performance of all the baselines consistently.  In few cases, we observe that MAE values of the distilled version of a model using even lesser training data is better than vanilla version of the same model using more training data and highlight it with *.}
  \label{tbl-limited-data}
\end{table*}
\section{Additional Experimental Results}
\subsection{Evaluation on Electronic Properties.}
Properties of crystalline materials can be broadly classified into two categories : 1) properties like formation energy, bandgap, total energy, bulk modulus, etc. which depend greatly on crystal structures and atom features, and 2) properties like $\epsilon_x$, $\epsilon_y$, $\epsilon_z$, n-Seebeck, n-PF, etc. which depend on the precise description of the materials’ electronic structure. We discuss performance of SOTA models with distilled knowledge from \our{} for the first class of properties in {\bf ``Downstream Task Evaluation"} section and report the MAE in Table \ref{tbl-full-data-1}. We further evaluate the effectiveness of CrysGNN for the second class of properties for both the vanilla and distilled model and report the MAE in Table \ref{tbl-full-data-2}. We observe, for this class of properties all the SOTA models had a higher MAE and though pre-training enhances the performance, the improvement is modest. In specific, average improvement in CGCNN, CrysXPP, GATGNN and ALIGNN are  6\%, 4\%, 3.6\% and 0.8\%, respectively. A potential reason is, electronic dielectric constant, Seebeck coefficients, and power factors all depend greatly on the precise description of the materials' electronic structure, which is neither captured by the SOTA models nor by the pre-trained \our{} framework explicitly. Hence error is high for these properties by SOTA models and injecting distilled structural information from the pre-trained model is able to achieve only modest improvements.\\
% On the contrary, for other properties (like $\epsilon_x$, $\epsilon_y$, $\epsilon_z$, n-Seebeck, n-PF, etc.) where SOTA models had higher MAE,  though pre-training enhances the performance, the improvement is modest. We report the MAEs for these properties for both the vanilla and distilled model in Table \ref{tbl-full-data-2} and average improvement in CGCNN, CrysXPP, GATGNN and ALIGNN are  6\%, 4\%, 3.6\% and 0.8\%, respectively. A potential reason is, electronic dielectric constant, Seebeck coefficients, and power factors all depend greatly on the precise description of the materials' electronic structure, which is neither captured by the SOTA models nor by the pre-trained \our{} framework explicitly. Hence error is high for these properties by SOTA models and injecting distilled structural information from the pre-trained model is not able to achieve much improvements.\\
\subsection{Effectiveness on sparse training dataset.}
A concise version of this result is presented in the main paper, we here elaborate some details.
To demonstrate the effectiveness of the pre-training in limited data settings, we conduct additional set of experiments under different training data split. In specific, we vary available training data from 20 to 60 \%, train different SOTA models and check their performance on test dataset. We report the MAE values of different baselines and their distilled version in Table \ref{tbl-limited-data} for five different properties, for which available data is very limited. We observe that the distilled version of any SOTA model consistently outperforms its vanilla version even in the limited training data setting, which illustrates the robustness of our pre-training framework. 
Specifically, for CGCNN and GATGNN, the improvements are more for 20\% training data, because these deep models suffer from data scarcity issue and  with distilled knowledge from the pre-trained model they are able to mitigate the issue. ALIGNN achieves moderate improvements as before but surprisingly, we found the least improvement (additional improvement for 20\% training data compared to other settings) for CrysXPP. CrysXPP itself pre-trains an autoencoder and transfers the learned information to the property predictor. Hence, the vanilla version performs well in the limited data settings.  \\
Moreover, in some cases, we observe that the MAE values of the distilled version of a model using lesser training data is better than vanilla version of the model using more training data. For example, while predicting Bandgap (MBJ), CGCNN (Distilled) with 20\% and 40\% training data outperforms CGCNN with 40\% and 60\% training data, respectively. These results verify that even with lesser training data, the distilled information from the large pre-trained model gives performance comparable to using larger training data by the original model.

\section{Ablation Study : Analysis of Different Pre-training Loss}
\begin{figure*}[ht]
	\centering
	\boxed{\includegraphics[width=2\columnwidth]{Figure/Legend.pdf}}
	\\
	\vspace*{-1mm}
	\subfloat[Shear Modulus (Gv)]{
		\boxed{\includegraphics[width=0.50\columnwidth, height=30mm]{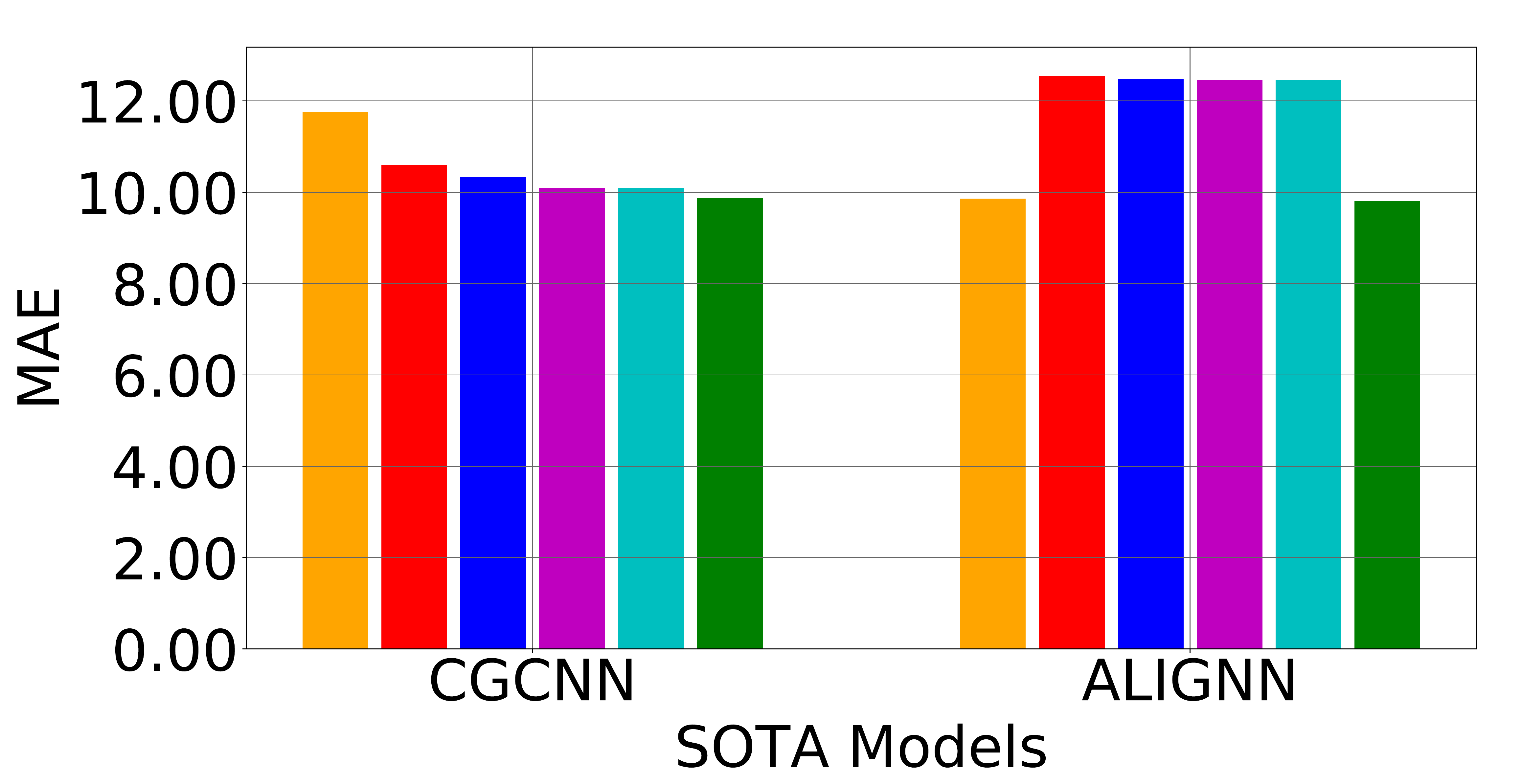}}}
	\subfloat[SLME(\%)]{
		\boxed{\includegraphics[width=0.50\columnwidth, height=30mm]{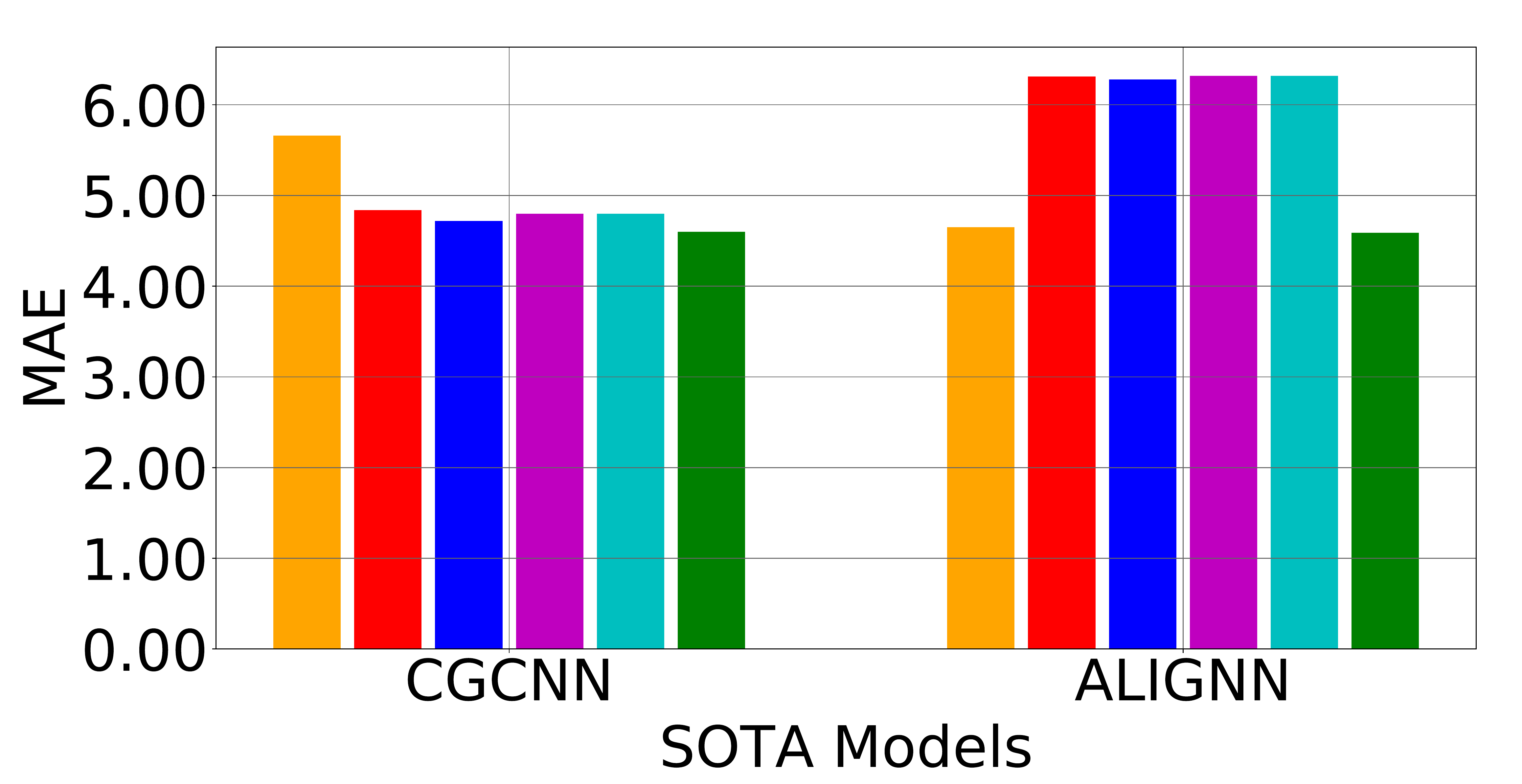}}}
	\subfloat[Spillage]{
		\boxed{\includegraphics[width=0.50\columnwidth, height=30mm]{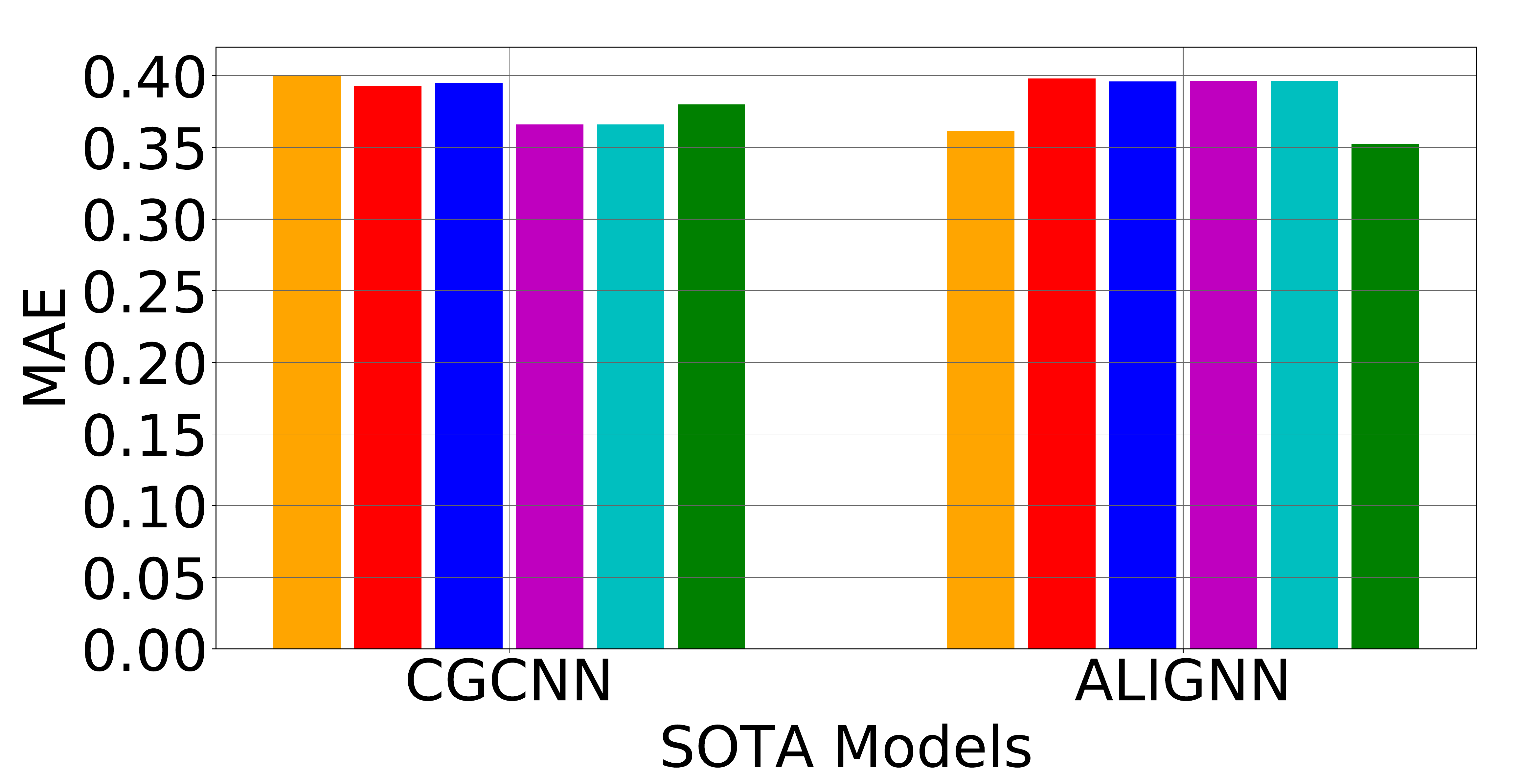}}}
	\subfloat[$\epsilon_x$ (MBJ)]{
		\boxed{\includegraphics[width=0.50\columnwidth, height=30mm]{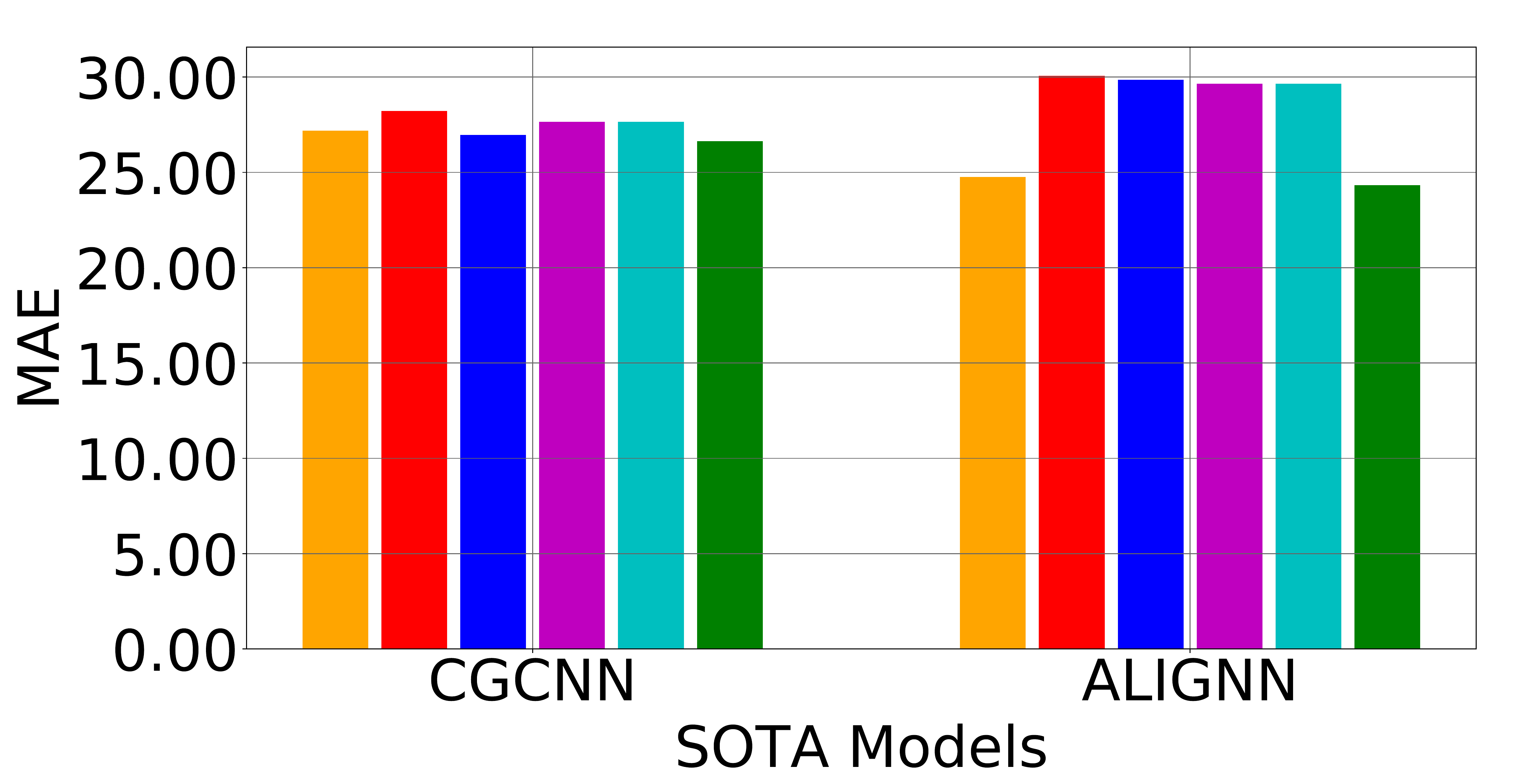}}}\\
	\subfloat[$\epsilon_y$ (MBJ) ]{
		\boxed{\includegraphics[width=0.50\columnwidth, height=30mm]{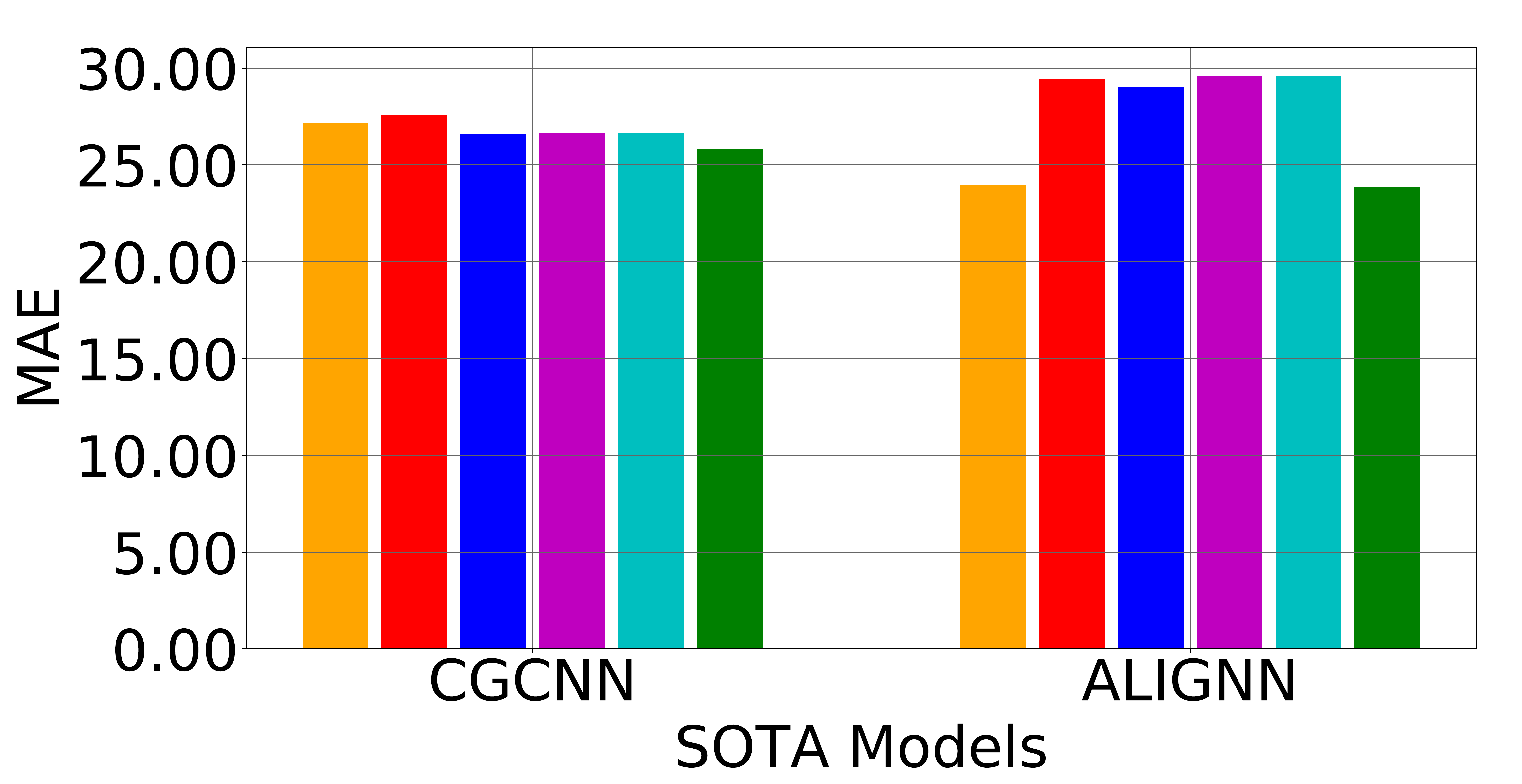}}}
	\subfloat[$\epsilon_z$ (MBJ) ]{
		\boxed{\includegraphics[width=0.50\columnwidth, height=30mm]{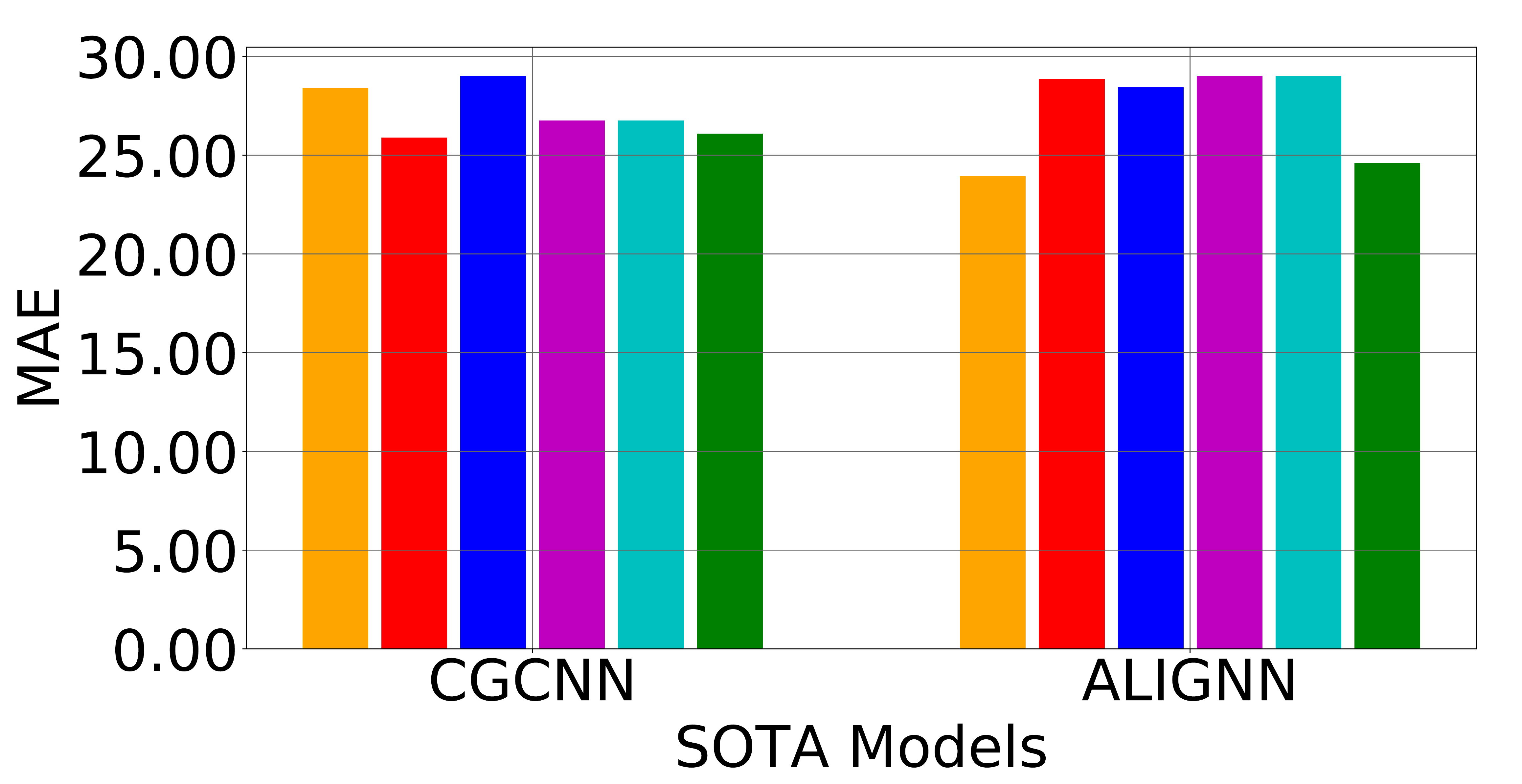}}}
% 	\subfloat[$\epsilon_x$ (OPT)]{
% 		\boxed{\includegraphics[width=0.50\columnwidth, height=30mm]{Figure/epsx.pdf}}}
% 	\subfloat[$\epsilon_y$ (OPT) ]{
% 		\boxed{\includegraphics[width=0.50\columnwidth, height=30mm]{Figure/epsy.pdf}}}\\
% 	\subfloat[$\epsilon_z$ (OPT) ]{
% 		\boxed{\includegraphics[width=0.50\columnwidth, height=30mm]{Figure/epsz.pdf}}}
	\subfloat[n-Seebeck]{
		\boxed{\includegraphics[width=0.50\columnwidth, height=30mm]{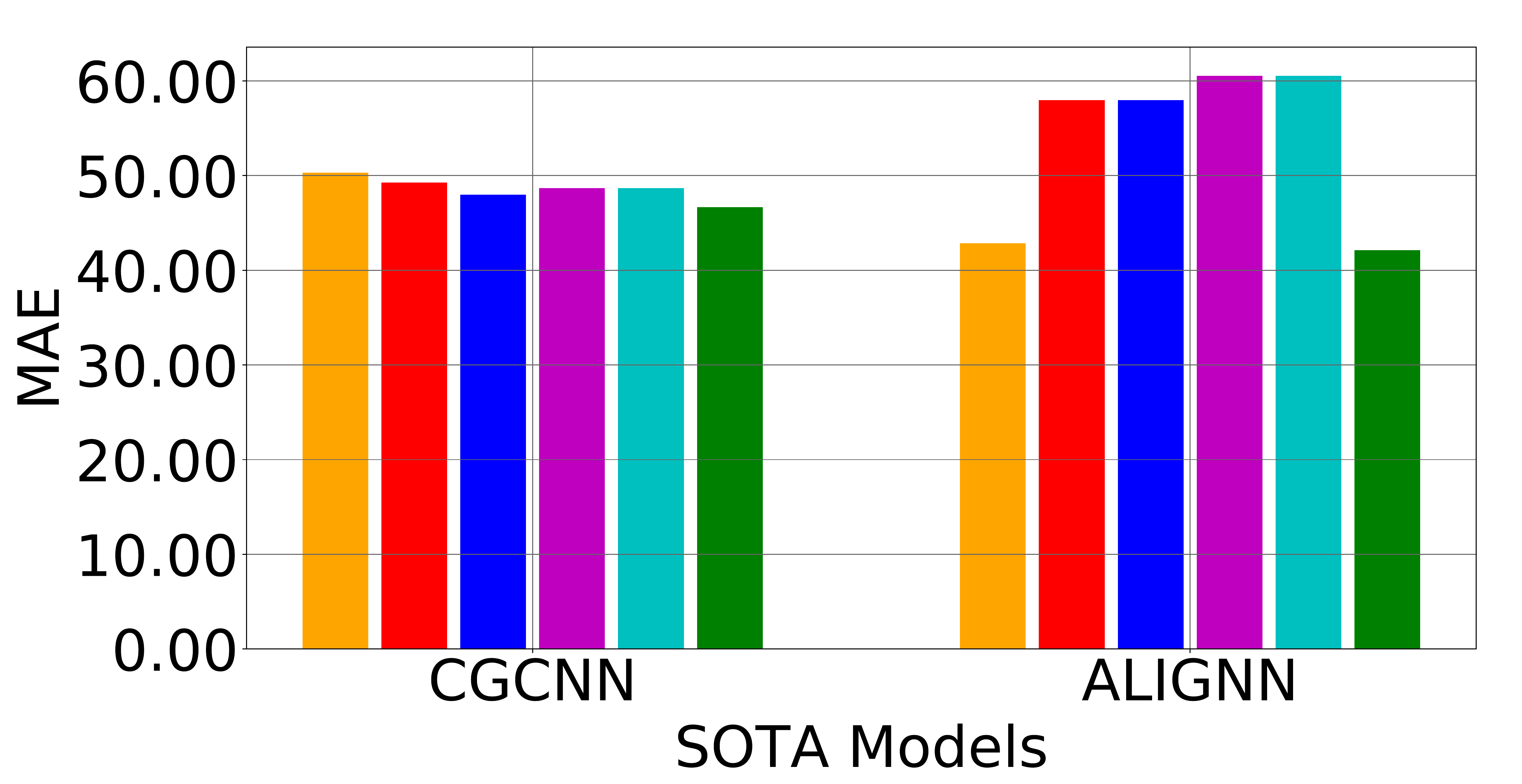}}}
	\subfloat[n-PF]{
		\boxed{\includegraphics[width=0.50\columnwidth, height=30mm]{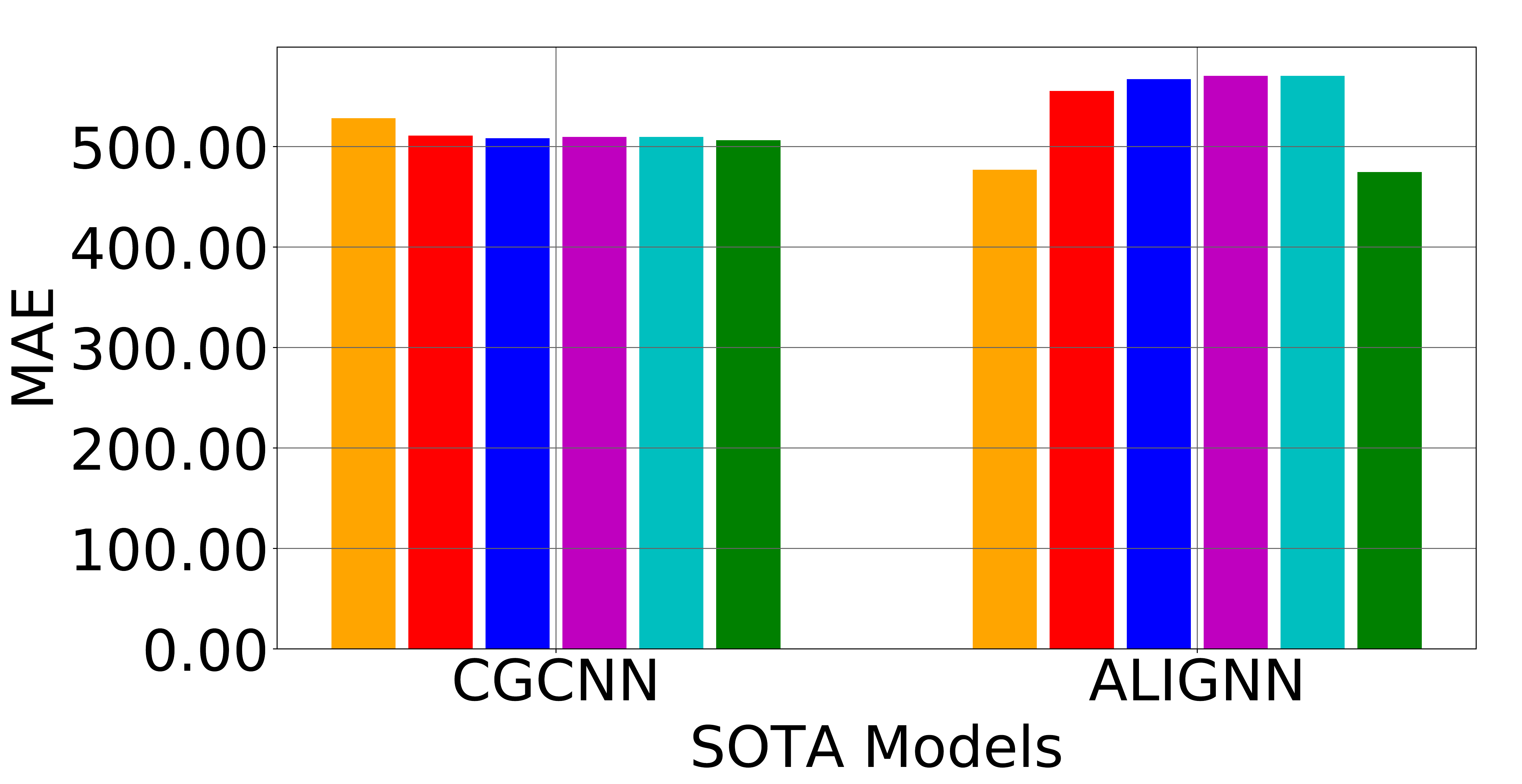}}}\\
	\subfloat[p-Seebeck]{
		\boxed{\includegraphics[width=0.50\columnwidth, height=30mm]{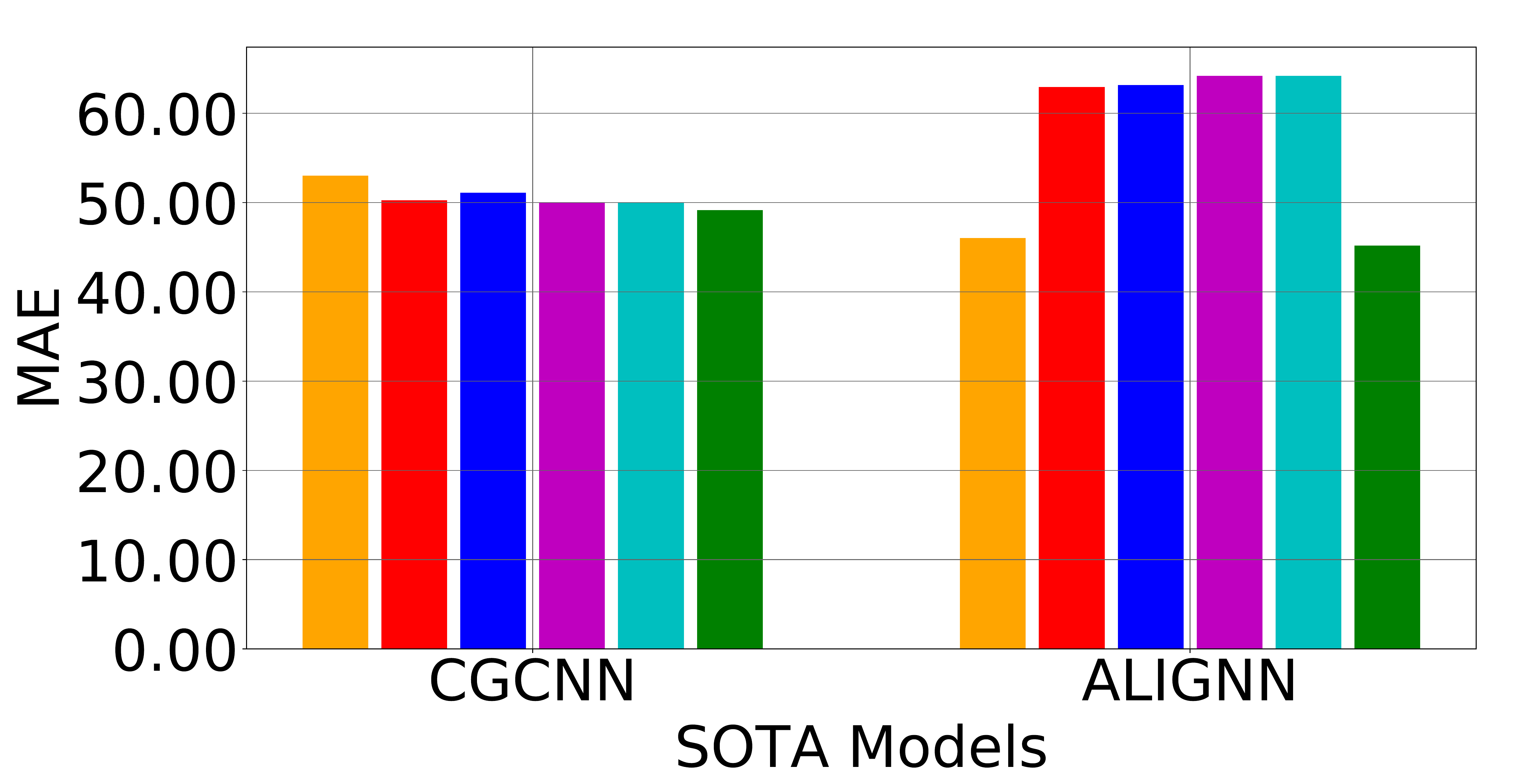}}}
	\subfloat[p-PF]{
		\boxed{\includegraphics[width=0.50\columnwidth, height=30mm]{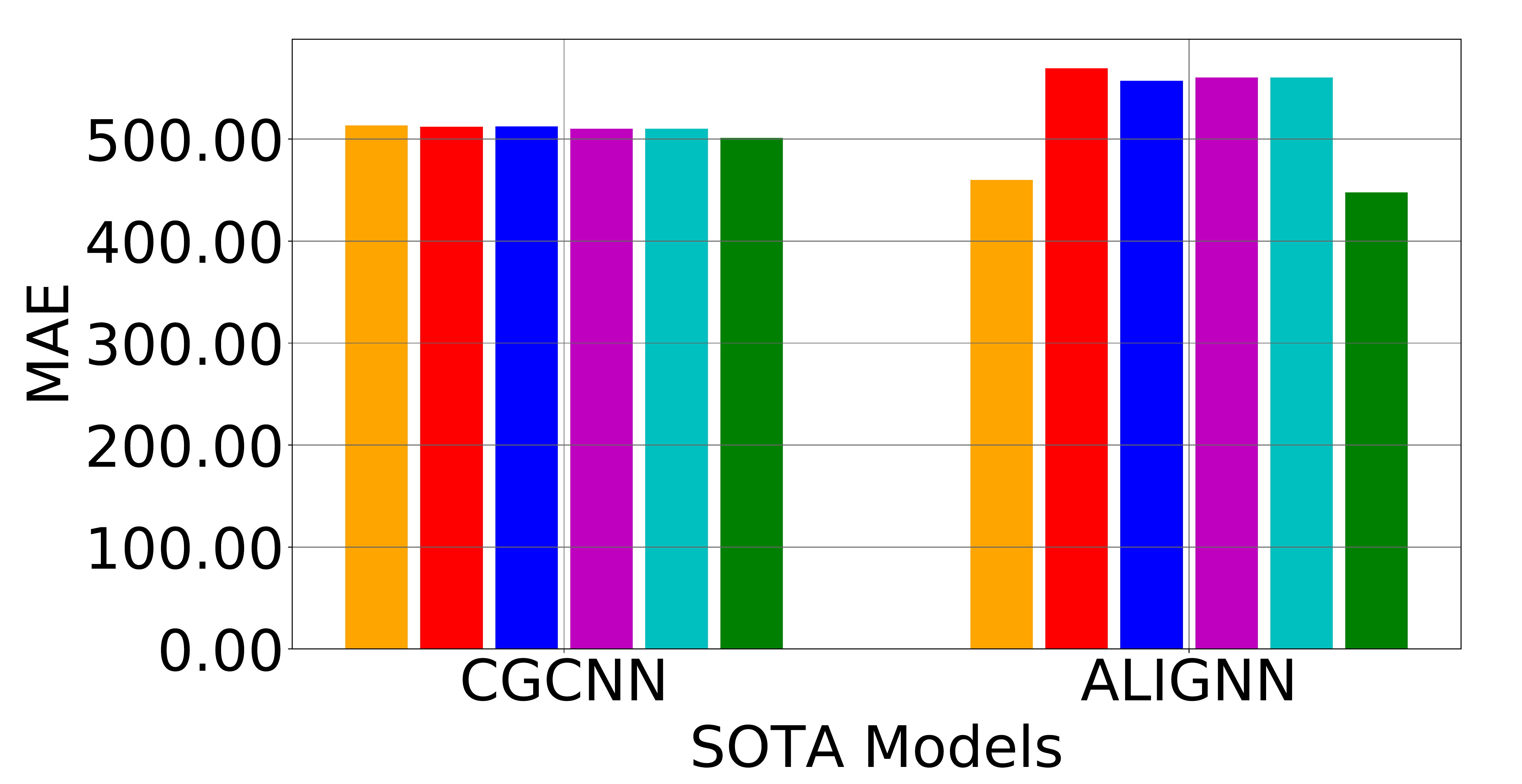}}}
	
	\caption{Summary of experiments of ablation study on importance of different pre-training loss components on \our{} training and eventually its effect on CGCNN and ALIGNN models on ten different properties from JARVIS DFT dataset(MAE for property prediction). (i) Vanilla: SOTA based model (without distillation) and all the other cases are SOTA models (distilled) from different pre-trained version of \our{}.
	(ii) Node : only node-level pre-training, (iii) Graph : only graph-level pre-training, (iv) Node + L(SG) : node-level and $\mathcal{L}_{SG}$, (v) Node + L(NTXent) : node-level and $\mathcal{L}_{NTXent}$ and (vi) \our{}~: both node and graph-level pre-training.}
	\label{fig:ablation_diff_loss_appendix}
\end{figure*}

A concise version of this result is presented in the main paper, we here elaborate some details.
We perform an ablation study to investigate the influence of different pre-training losses in enhancing the SOTA model performance. While pre-training \our{} (Eq. \ref{eq:pretrain_loss}), we capture both local chemical and global structural information via node and graph-level decoding, respectively. 
Further, we are curious to know the influence of each of these decoding policies independently in the downstream property prediction task.  
\tblak{In specific, we conduct the ablation experiments, where we pre-train \our{} with (a) only node-level decoding ($\mathcal{L}_{FR}$, $\mathcal{L}_{CR}$), (b) only graph-level decoding ($\mathcal{L}_{SG}$, $\mathcal{L}_{NTXent}$). Further, we perform ablations with individual graph-level losses, and pretrain with (c) removing $\mathcal{L}_{NTXent}$ (node-level with $\mathcal{L}_{SG}$ (space group)) and (d) removing $\mathcal{L}_{SG}$ (node-level with $\mathcal{L}_{NTXent}$(crystal system)).} We train two baseline models, CGCNN and ALIGNN, with distilled knowledge from all the aforementioned variants of the pre-trained model and evaluate the performance on four crystal properties. We report performance of both the experiments for four properties in Fig.\ref{fig:ablation_diff_loss} and for ten different properties in Fig.\ref{fig:ablation_diff_loss_appendix}.\\
We can observe clearly that all the variants offer significant performance gain in all four properties using the combined node and graph-level pre-training, compared to node-level or graph-level pre-training separately. Only exception is formation energy, where only node-level pre-training produces less error compared to other variants, in both the baseline. Formation energy of a crystal is defined as the difference between the energy of a unit cell comprised of $N$ chemical species and the sum of the chemical potentials of all the $N$ chemical species. Hence pre-training at the node-level (node features and connection) is adequate for enhancing performance of formation energy prediction and incorporating graph-level information works as a noisy information, which degrades the performance. \tblak{We also observe improvement in performance using both supervised and contrastive graph-level losses ($\mathcal{L}_{SG}$ and $\mathcal{L}_{NTXent}$), compared to using only one of them, which proves that the learned representation via supervised and contrastive learning is more expressive that using any one of them.}
Moreover, in ALIGNN, with either node or graph-level pre-training separately, performance degrades across different properties. ALIGNN explicitly captures the three body interactions which drive its performance, to replicate that inclusion of both node and graph information is necessary. 
 
\subsection{Limitations and Future Work.}
Though our proposed pre-trained model is able to enhance the performance of all the state of the art baseline models, improvements for all types of SOTA models are not consistent and we found lesser improvement in case of complex models like GATGNN and ALIGNN. This provides scope for further investigation on designing a more complex and deeper pre-trained model. There are also scopes of exploring different graph representations, loss variations, distillation strategies, etc. Of course, the very fact that the simple model, \our{}, can provide substantial improvements, lays the foundation  for such future exploration.  Moreover, in this present work, we have focused on predicting crystal properties, which is a graph level regression task. However, the same framework can be used to  explore the effect of pre-training on other graph level tasks (classification or clustering) or node level tasks which would be one of our future works.
\end{document}